\documentclass[10pt,twocolumn,letterpaper]{article}

\usepackage{iccv}
\usepackage{times}
\usepackage{epsfig}
\usepackage{graphicx}
\usepackage{amsmath}
\usepackage{amssymb}

\usepackage{url}
\usepackage{booktabs}       
\usepackage{amsfonts}       
\usepackage{nicefrac}       
\usepackage{microtype}      
\usepackage{enumitem} 
\usepackage{wrapfig}
\usepackage{xspace}
\usepackage{graphicx}
\usepackage{caption}
\usepackage{subcaption}
\usepackage{makecell}
\usepackage{comment}
\usepackage[ruled, vlined]{algorithm2e}
   \usepackage{cuted}

\usepackage{physics}
\usepackage{lipsum}
\usepackage{array}
\usepackage{amsmath}
\DeclareMathOperator*{\argmax}{arg\,max}

\usepackage{enumitem}

\usepackage{amssymb}
\usepackage{amsmath,amsfonts}
\usepackage{amsopn}
\usepackage{bm} 
\usepackage{multirow}

\newcommand{\vct}[1]{\boldsymbol{#1}} 

\newcommand{\ProbOpr}[1]{\mathbb{#1}}

\newcommand{\expect}[2]{%
\ifthenelse{\equal{#2}{}}{\ProbOpr{E}_{#1}}
{\ifthenelse{\equal{#1}{}}{\ProbOpr{E}\left[#2\right]}{\ProbOpr{E}_{#1}\left[#2\right]}}} 

\newcommand{\ve}{\vct{e}}

\newcommand{\vw}{\vct{w}}

\newcommand{\eat}[1]{}
\newcommand{\method}[1]{\textsc{#1}}

\newcommand{\mixup}{\method{MixUp}\xspace}

\usepackage[pagebackref=true,breaklinks=true,colorlinks,bookmarks=false]{hyperref}
\usepackage[nameinlink,capitalize]{cleveref}
\usepackage{xcolor}
\definecolor{lightpurple}{RGB}{178,145,226}
\definecolor{lightblue}{RGB}{0,128,255}
\definecolor{yellow}{rgb}{1,1,0}

\newcommand{\mist}{\method{MiST}\xspace}

\newcommand{\vmist}{\textbf{\method{MiST}}\xspace}
\newcommand{\code}{\method{DeCoTa}\xspace}
\newcommand{\vcode}{\textbf{\method{DeCoTa}}\xspace}

\newcommand{\DS}{D_{\mathcal{S}}{}}
\newcommand{\DT}{D_{\mathcal{T}}{}}
\newcommand{\DU}{D_{\mathcal{U}}{}}

\renewcommand{\paragraph}[1]{\vspace{-0.5ex}\textbf{#1}}

\iccvfinalcopy

\def\iccvPaperID{10085}

\ificcvfinal\fi

\begin{document}

\title{Deep Co-Training with Task Decomposition \\for Semi-Supervised Domain Adaptation}

\author{
$\textbf{Luyu Yang}^{1}, \textbf{Yan Wang}^{2}, \textbf{Mingfei Gao}^{3}$,\\[0.2em]
$\textbf{Abhinav Shrivastava}^{1}, \textbf{Kilian Q. Weinberger}^{2}, \textbf{Wei-Lun Chao}^{4}, \textbf{Ser-Nam Lim}^{5}$
\\[0.5em]
\medskip
$^{1}\text{University of Maryland}$\quad
$^{2}\text{Cornell University}$\quad
$^{3}\text{Salesforce Research}$\quad\\
\smallskip
$^{4}\text{Ohio State University}$\quad
$^{5}\text{Facebook AI}$
\vspace{-0.1in}
}

\maketitle
\ificcvfinal\thispagestyle{empty}\fi

\begin{abstract}

Semi-supervised domain adaptation (SSDA) aims to adapt models trained from a labeled source domain to a different but related target domain, from which unlabeled data and a small set of labeled data are provided. Current methods that treat source and target supervision without distinction overlook their inherent discrepancy, resulting in a source-dominated model that has not effectively use the target supervision. In this paper, we argue that the labeled target data needs to be distinguished for effective SSDA, and propose to explicitly decompose the SSDA task into two sub-tasks: a semi-supervised learning (SSL) task in the target domain and an unsupervised domain adaptation (UDA) task across domains. 
By doing so, the two sub-tasks can better leverage the corresponding supervision and thus yield very different classifiers. To integrate the strengths of the two classifiers, we apply the well established co-training framework, in which the two classifiers exchange their high confident predictions to iteratively ``teach each other'' so that both classifiers can excel in the target domain.
We call our approach \textbf{De}ep \textbf{Co}-training with \textbf{Ta}sk decomposition (\vcode).
\code requires no adversarial training and is easy to implement.
Moreover, \code is well founded on the theoretical condition of when co-training would succeed. As a result, \code achieves state-of-the-art results on several SSDA datasets, outperforming the prior art by a notable $4\%$ margin on DomainNet. \textit{Code is available at \url{https://github.com/LoyoYang/DeCoTa}}.
\end{abstract}

\section{Introduction}
\label{s_intro}

\begin{figure}[h]
	\centering
    \includegraphics[width=0.95\linewidth]{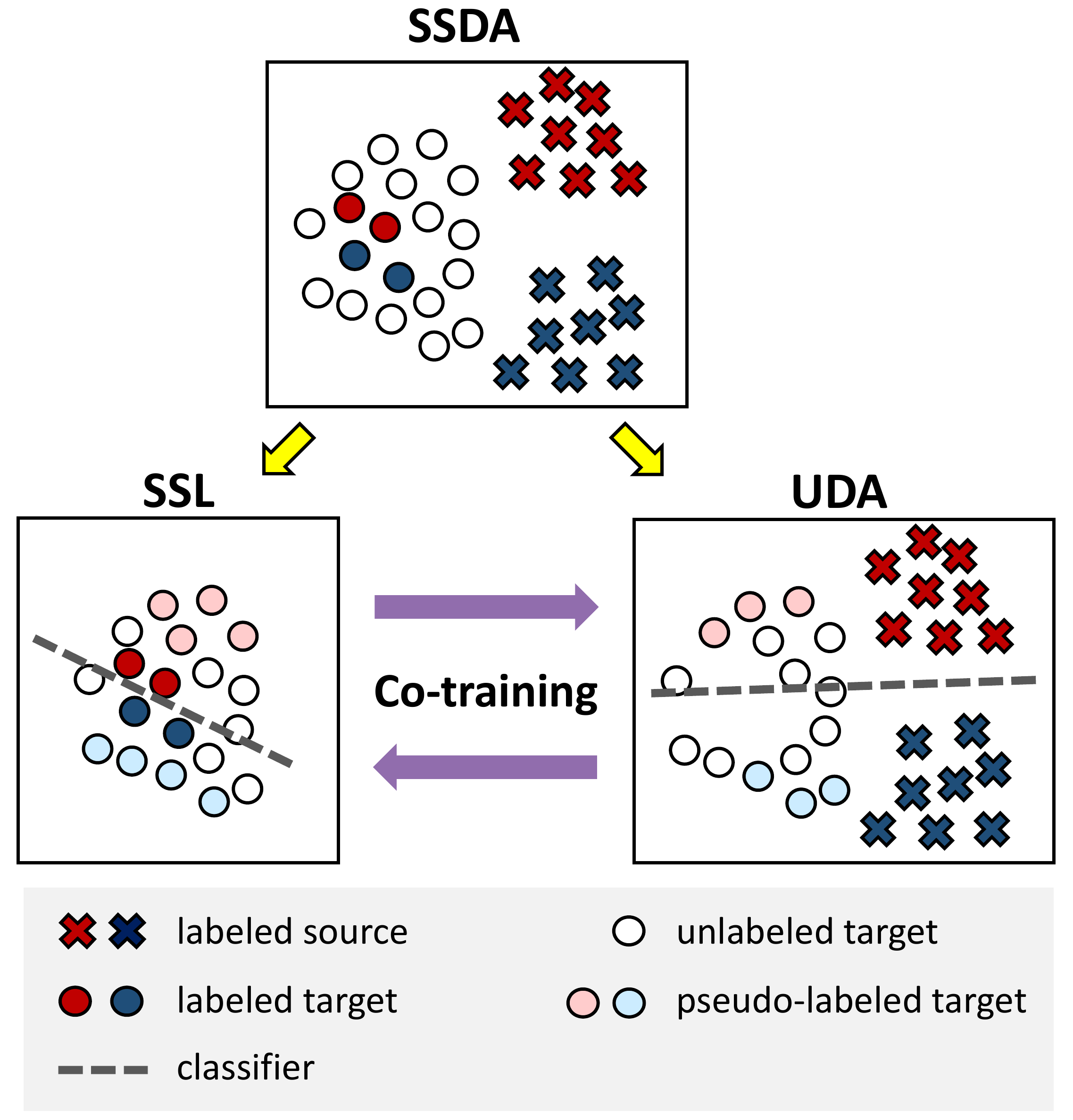}
    \vspace{-10pt}
    \caption{\small \textbf{De}ep \textbf{Co}-training with \textbf{Ta}sk decomposition (\vcode). We decompose semi-supervised domain adaptation (SSDA) into two sub-tasks: semi-supervised learning (SSL) in the target domain, and unsupervised DA (UDA) across domains. The two sub-tasks offer different pseudo-label confidences to the unlabeled data (light blue \& light red circles), which we leverage via co-training: exchanging their high confident predictions to teach each other.}
    \label{fig:main}
    \vspace{-10pt}
\end{figure}

Domain adaptation (DA) aims to adapt machine learned models from a source domain to a related but different target domain \cite{ben2010theory,gong2014learning,saito2018maximum,ganin2016domain}. DA is particularly important in settings where labeled target data is hard to obtain, but labeled source data is plentiful \cite{tsai2018learning,peng2019moment,hoffman2017cycada}, \eg, adaptation from synthetic to real images \cite{hoffman2017cycada,saleh2018effective,ros2016synthia,richter2016playing,sankaranarayanan2018learning} and adaptation to a new or rare environment \cite{chen2017no,yan2020domain,sakaridis2018semantic,chen2018domain}.
Most of the existing works focus on the unsupervised domain adaptation (UDA) setting, 
in which the target domain is completely unlabeled. Several recent works, however, show that adding merely a tiny amount of target labeled data (\eg, just one labeled image per class) can notably boost the performance \cite{saito2019semi,kim2020attract,qin2020opposite,ao2017fast,li2018semi,li2020online,donahue2013semi,yao2015semi}, suggesting that this setting may be more promising for domain adaptation to succeed.
In this paper, we thus focus on the latter setting, which is referred to as semi-supervised domain adaptation (SSDA).

Despite the seemingly nuanced difference between the two settings, methods that are effective for SSDA and UDA can vary substantially. For instance, \cite{saito2019semi} showed that directly combining the labeled source and labeled target data and 
then applying popular UDA algorithms like domain adversarial learning \cite{ganin2016domain} or entropy minimization \cite{grandvalet2005semi} can hardly improve the performance. In other words, the labeled target data have not been effectively used. Existing methods \cite{saito2019semi,qin2020opposite,kim2020attract} therefore propose additional objectives to strengthen the influence of labeled target data in SSDA.

Intrigued by these findings, we investigate the characteristics of SSDA further and emphasize two fundamental challenges.
First, the amount of labeled source data is much larger than that of labeled target data. Second, the two data are inherently different in their distributions. \textbf{A single classifier} learned together with both sources of supervision is thus easily dominated by the labeled source data and is unable to take advantage of the additional labeled target data. 

To resolve this issue, we propose to explicitly decompose the two sources of supervision and learn \textbf{two distinct classifiers} whose goals are however shared: to classify well on the unlabeled target data. To this end, we pair the labeled source data and the unlabeled target data to learn one classifier, which is essentially a UDA task. For the other classifier, we pair the labeled and unlabeled target data, which is essentially a semi-supervised learning (SSL) task. That is, \textbf{we explicitly decompose SSDA into two well-studied tasks.} 

For each sub-task, one may apply any existing algorithms independently. In this paper, we however investigate the idea of learning the two classifiers jointly
for two compelling reasons. First, the two tasks share the same goal and same unlabeled data, meaning that they are \emph{correlated}. Second, learning with distinct labeled data implies that the two classifiers will converge differently in what types of mistakes they make and on which samples they are confident and correct, meaning that they are \emph{complementary} to each other.

We therefore propose to learn the two classifiers jointly via co-training \cite{blum1998combining,balcan2005co,chen2011automatic}\footnote{We note that, co-training \cite{blum1998combining} and co-teaching \cite{han2018co} share similar concepts but are fundamentally different. See \ref{s_related} for a discussion.}, which is arguably one of the most established algorithm for learning with multi views: in our case, two correlating and complementary tasks.
The approach is straightforward: train a separate classifier on each task using its labeled data, and use them to create pseudo-labels for the unlabeled data. As the two classifiers are trained with distinct supervision, they will yield different predictions. In particular, there will be samples that only one classifier is confident about (and more likely to be correct). By labeling these samples with the confident classifier’s predictions and adding
them to the training set of the other classifier to re-train on, the two classifiers are essentially “teaching each other” to improve. To this end, we employ a simple \emph{pseudo-labeling-based algorithm with deep learning}, similar to \cite{berthelot2019mixmatch}, to train each classifier. Pseudo-labeling-based algorithms have been shown powerful for both the UDA and SSL tasks \cite{wei2020theoretical,kumar2020understanding}. In other words, we can apply the same algorithm for both sub-tasks, greatly simplifying our overall framework which we name \vcode: \textbf{De}ep \textbf{Co}-training with \textbf{Ta}sk Decomposition (\cref{fig:main} gives an illustration).

We evaluate \code on two benchmark datasets for SSDA: DomainNet \cite{peng2019moment} and Office-home \cite{venkateswara2017deep}. While very simple to implement and without any adversarial training \cite{saito2019semi,qin2020opposite}, \code significantly outperforms the state-of-the-art results \cite{qin2020opposite,kim2020attract} on DomainNet by over $4\%$ and is on a par with them on Office-home. \emph{We attribute this to the empirical evidence that 
our task decomposition fits the theoretical condition of relaxed $\epsilon$-expandability~\cite{chen2011automatic,balcan2005co}, which is sufficient for co-training to succeed.}
Another strength of \code is that it requires no extra learning process like feature decomposition to create views from data~\cite{chen2011automatic,Qiao2018DeepCF,chen2011co}. To the best of our knowledge, our paper is the first to enable deep learning with co-training on SSDA.

The contributions of this work are as follow. (1) We explicitly decompose the two very different sources of supervision, labeled source and labeled target data, in SSDA. (2) We present \code, a simple deep learning based co-training approach for SSDA to jointly learn two classifiers, one for each supervision. (3) we provide intermediate results and insights that illustrate why \code works. Specifically, we show that \code satisfies the $\epsilon$-expandability requirement \cite{balcan2005co} of co-training.
(4) Lastly, we support this work with strong empirical results that outperform state-of-the-art.
\section{Related Work}
\label{s_related}

\noindent\textbf{Unsupervised domain adaptation (UDA).} UDA has been studied extensively. 
Many methods \cite{long2015learning, shu2018dirt, tzeng2014deep} matched the feature distributions between domains by minimizing their divergence.
One mainstream approach is by domain adversarial learning \cite{ganin2016domain,hoffman2017cycada,wang2019transferable,pei2018multi,volpi2018adversarial,xu2020adversarial,yan2020domain, yang2020curriculum}.
More recent works \cite{saito2017adversarial,saito2018maximum,lee2019drop,shu2018dirt} learn features based on the cluster assumption \cite{grandvalet2005semi}: classifier boundaries should not cross high density
target data regions. For example, \cite{saito2017adversarial,saito2018maximum} attempted to push target features away from the boundary, using minmax training.
Some other approaches employ self-training with pseudo-labeling \cite{lee2013pseudo,mcclosky2006effective,mcclosky2006reranking,banko2001scaling} to progressively label unlabeled data and use them to fine-tune the model~\cite{chen2011co,kim2019self,zou2018unsupervised,khodabandeh2019robust,tao2018zero,liang2019distant,inoue2018cross,kumar2020understanding}.
A few recent methods use \mixup~\cite{zhang2017mixup}, but mainly to augment adversarial learning based UDA approaches  (\eg, \cite{ganin2016domain}) by stabilizing the domain discriminator~\cite{tang2020adversarial,xu2020adversarial} or smoothing the predictions~\cite{mao2019virtual,yan2020improve}.
In contrast, we apply \mixup to create better pseudo-labeled data for co-training, without adversarial learning.

\noindent\textbf{Semi-supervised domain learning (SSDA).}
SSDA attracts less attention in DA, despite its promising scenario in balancing accuracy and labeling effort. With few labeled target data, SSDA can quickly reshape the class boundaries to boost the accuracy \cite{saito2019semi,qin2020opposite}. Many SSDA works are proposed prior to deep learning~\cite{yao2015semi,li2018semi,hoffman2013efficient,pereira2018semi}, matching features while maintaining accuracy on labeled target data. \cite{ao2017fast,tzeng2015simultaneous} employed knowledge distillation \cite{hinton2015distilling} to regularize the training on labeled target data. More recent works use deep learning, and find that the popular UDA principle of aligning feature distributions could fail to learn discriminative class boundaries in SSDA \cite{saito2019semi}. \cite{saito2019semi} thus proposed to gradually move the class prototypes (used to derive class boundaries) to the target domain in a minimax fashion;
\cite{qin2020opposite} introduced opposite structure learning to cluster target data and scatter source data to smooth the process of learning class boundaries. Both works \cite{qin2020opposite,saito2019semi} and \cite{kim2020attract} concatenate the target labeled data with the source data to expand the labeled data. \cite{li2020online} incorporates meta-learning to search for better initial condition in domain adaptation. SSDA is also related to~\cite{su2020active,prabhu2020active}, in which active learning is incorporated to label data for improving domain adaptation.

\noindent\textbf{Co-training.} 
Co-training, a powerful semi-supervised learning (SSL) method proposed in \cite{blum1998combining},
looks at the available data with two views 
from which two models are trained interactively.
By adding the confident predictions of one model to the training set of the other, co-training enables the models to ``teach each other''. There were several assumptions 
to ensure co-training's effectiveness~\cite{blum1998combining}, which were later relaxed by~\cite{balcan2005co} with the notion of $\epsilon$-expandability. 
\cite{chen2011automatic} broadened the scope of co-training to a single-view setting by learning to decompose a fixed feature representation into two artificially created views; \cite{chen2011co} subsequently extended this framework to use co-training for (semi-supervised) domain adaptation\footnote{Similar to \cite{qin2020opposite,saito2019semi}, \cite{chen2011co} simply concatenated the target labeled data with the source data to expand the labeled data.}.
A recent work~\cite{Qiao2018DeepCF} extended co-training to deep learning models, by encouraging two models to learn different features and behave differently on single-view data. 
One novelty of \code is that it works with single-view data (both the UDA and SSL tasks are looking at images) but requires no extra learning process like feature decomposition to {artificially create} views from such data \cite{chen2011automatic,Qiao2018DeepCF,chen2011co}. 

\noindent\textbf{Co-training vs. co-teaching.}
Co-teaching \cite{han2018co} was proposed for learning with noisy data, which shares a similar procedure to co-training by learning two models to filter out noisy data for each other. There are several key differences between them and \emph{\vcode
is based on co-training}. As in~\cite{han2018co}, co-teaching is designed for supervised learning with noisy labels, while co-training is for learning
with unlabeled data by leveraging two views. \code decomposes SSDA into two tasks (two views) to leverage their difference to improve the performance — the core concept of co-training~\cite{chen2011co}. In contrast, co-teaching does not need two views.
Further, co-teaching relies on the memorization of neural nets to select small loss samples to teach the other classifiers, while \code selects high confident ones from unlabeled data.
\section{\underline{De}ep \underline{Co}-training with \underline{Ta}sk Decomposition}
\label{s_approach}

\subsection{Approach Overview}
\label{ss_code}

Co-training strategies have traditionally been applied to data with two views, e.g., audio and video, or webpages with HTML source and link-graph, after which a classifier is trained in each view and they teach each other on the unlabeled data.
This is the original formulation from Blum and Mitchell~\cite{blum1998combining},
which is later extended to single-view data by~\cite{chen2011automatic} for linear models and by~\cite{Qiao2018DeepCF} for deep neural networks. Both methods require additional objective functions or tasks (\eg, via generating adversarial examples~\cite{goodfellow2014explaining}) to learn to create \emph{artificial} views such that co-training can be applied.

In this paper, we have however discovered that in semi-supervised domain adaptation (SSDA), one can actually conduct co-training using single-view data (all are images) without such an additional learning subroutine. The key is to leverage the inherent discrepancy of the labeled data (\ie, supervision) provided in SSDA: the labeled data from the source domain, $\DS=\{(s_i,y_i)\}_{i=1}^{N_S}$, and the labeled data from the target domain, $\DT=\{(t_i,y_i)\}_{i=1}^{N_T}$, which is 
usually much smaller than $\DS$.
By combining each of them with the unlabeled samples from the target domain, $\DU=\{u_i\}_{i=1}^{N_U}$, we can construct two sub-tasks in SSDA: 
\begin{itemize} [leftmargin=*,itemsep=1pt,topsep=2pt]
  \item an \textbf{unsupervised domain adaptation (UDA)} task that trains a model \underline{$\vw_{\color{blue}g}$ using $\DS$ and $\DU$},
  \item a \textbf{semi-supervised learning (SSL)} task that trains another model \underline{$\vw_{\color{red}f}$ using $\DT$ and $\DU$}. 
\end{itemize}

We learn both models by mini-batch stochastic gradient descent (SGD). At every iteration, we sample three data sets, $S=\{(s_b, y_b)\}_{b=1}^B$ from $\DS$, $T=\{(t_b, y_b)\}_{b=1}^B$ from $\DT$, and $U=\{u_b\}_{b=1}^B$ from $\DU$, where $B$ is the mini-batch size. We can then predict on $U$ using the the two models $\vw_{\color{blue}g}$ and $\vw_{\color{red}f}$, creating the pseudo-label sets $U^{({\color{red}f})}$ and $U^{({\color{blue}g})}$ that will be used to update $\vw_{\color{red}f}$ and $\vw_{\color{blue}g}$,
\begin{align}
    U^{({\color{red}f})} = &\{(u_b, \hat{y_b}= \argmax_c p(c|u_b; \vw_{\color{blue}g})); \nonumber\\
    &\text{ if } \max_c p(c|u_b; \vw_{\color{blue}g}) > \tau \}, \nonumber\\
    U^{(\color{blue}g)} = &\{(u_b, \hat{y_b}= \argmax_c p(c|u_b; \vw_{{\color{red}f}})); \nonumber\\
    &\text{ if } \max_c p(c|u_b; \vw_{{\color{red}f}}) > \tau \},
\label{eq_code}
\end{align} 
where $u_b$ is an unlabeled sample drawn from $U$, $p(c|u_b; \cdot)$ is the predicted probability for a class $c$, and $\tau$ is the threshold for pseudo-label selection. In other words, we use one model's (say $\vw_{\color{blue}g}$) high confident prediction to create pseudo-labels for $u_b$, which is then included in $U^{({\color{red}f})}$ that will be used to train the other model $\vw_{{\color{red}f}}$.
By looking at $U^{({\color{red}f})}$ and $U^{({\color{blue}g})}$ jointly, we are indeed asking one model to simultaneously be a \emph{teacher} and a \emph{student}: it provides confident pseudo-labels for the other model to learn from, and learns from the other model's confident pseudo-labels.

We call this approach \vcode, which stands for \underline{De}ep \underline{Co}-training with \underline{Ta}sk Decomposition. In the following, we will discuss how to improve the pseudo-label quality (\ie, its coverage and accuracy) for \vcode , and provide in-depth analysis why \code works.

\begin{center}
\begin{algorithm}[t]
\SetAlgoLined
\small
\caption{The \vcode algorithm}
\SetKwInOut{Input}{Input}
\SetKwInOut{Output}{Output}
\Input{$\vw_{\color{red}f}$ and $\vw_{\color{blue}g}$, learning rate $\eta$, batch size $B$, iteration $N_{\max}$, beta distribution coefficient $\alpha$, confidence threshold $\tau$, data $\DS$, $\DT$, $\DU$;}
\For{$n\leftarrow 1$ \KwTo $N_{\max}$}{
{\textbf{Sample} $S=\{(s_b, y_b)\}_{b=1}^B$ from $\DS$, \\
\textbf{Sample} $T=\{(t_b, y_b)\}_{b=1}^B$ from $\DT$, \\
\textbf{Sample}  $U=\{u_b\}_{b=1}^B$ from $\DU$;}\\
{\textbf{Set} $U^{({\color{red}f})}=\emptyset$, $U^{({\color{blue}g})}=\emptyset$;}\\
\For{$b\leftarrow 1$ \KwTo $B$}{
\If{$\max_c p(c|u_b; \vw_{\color{blue}g})> \tau$}{
    \textbf{Update} $U^{({\color{red}f})}\leftarrow U^{({\color{red}f})} + \{(u_b, \hat{y}_b)\}$, $\hat{y}_b = \argmax_c p(c|u_b; \vw_{\color{blue}g})$;
    
}
\If{$\max_c p(c|u_b; \vw_{\color{red}f})> \tau$}{
    \textbf{Update} $U^{({\color{blue}g})}\leftarrow U^{({\color{blue}g})} + \{(u_b, \hat{y}_b)\},$ $\hat{y}_b = \argmax_c p(c|u_b; \vw_{\color{red}f})$;
}
}
\textbf{Obtain} $\tilde{U}^{({\color{red}f})} = \{\text{\mixup}({U}^{({\color{red}f})}_i, T_i; \alpha)\}_{i=1}^{|{U}^{({\color{red}f})}|}$;\\
\textbf{Obtain} $\tilde{U}^{({\color{blue}g})} = \{\text{\mixup}({U}^{({\color{blue}g})}_i, S_i; \alpha)\}_{i=1}^{|{U}^{({\color{blue}g})}|}$;\\
\textbf{Update} $\vw_{\color{red}f} \leftarrow \vw_{\color{red}f} -  \eta \left(\nabla\mathcal{L}(\vw_{\color{red}f}, T) + \nabla\mathcal{L}(\vw_{\color{red}f}, \tilde{U}^{({\color{red}f})})\right)$;\\
\textbf{Update} $\vw_{\color{blue}g} \leftarrow \vw_{\color{blue}g} -  \eta \left(\nabla\mathcal{L}(\vw_{\color{blue}g}, S)  + \nabla\mathcal{L}(\vw_{\color{blue}g}, \tilde{U}^{({\color{blue}g})})\right)$;\\
}
\Output{$\vw_f$ and $\vw_g$ (for model ensemble).}
\label{alg_code}
\end{algorithm}
\vspace{-0.35in}
\end{center}

\subsubsection{\code with High-quality Pseudo-labels}
\label{ss_mixup}

The pseudo-labels acquired from each model are understandably noisy. At the beginning of the training, this problem is especially acute, and affects the efficacy of the model as the training progresses. Our experience shows that mitigation is necessary to handle noise in the pseudo-labels to further enhance \vcode, for which we follow recent works of SSL \cite{berthelot2019mixmatch} to apply \mixup \cite{zhang2017mixup,mai2019metamixup}. \mixup is an operation to construct virtual examples by convex combinations. Given two labeled examples $(x_1, y_1)$ and $(x_2, y_2)$, we define $\mixup\left((x_1, y_1), (x_2, y_2); \alpha\right)$
\begin{align}
    &\lambda\sim  \text{Beta} (\alpha, \alpha), \hspace{20pt} \nonumber\\
    &\tilde{x} =  (1-\lambda) x_1 + \lambda x_2,\hspace{20pt} \tilde{y} =  (1-\lambda) \ve_{y_1} + \lambda \ve_{y_2} \label{eq_mixup}
\end{align}
to obtain a virtual example $(\tilde{x}, \tilde{y})$, where $\ve_{y}$ is a one-hot vector with the $y^{th}$ element being 1. $\lambda$ controls the degree of \mixup while Beta refers to the standard beta distribution.

\begin{figure}[t]
	\centerline{
    \includegraphics[width=0.52\textwidth]{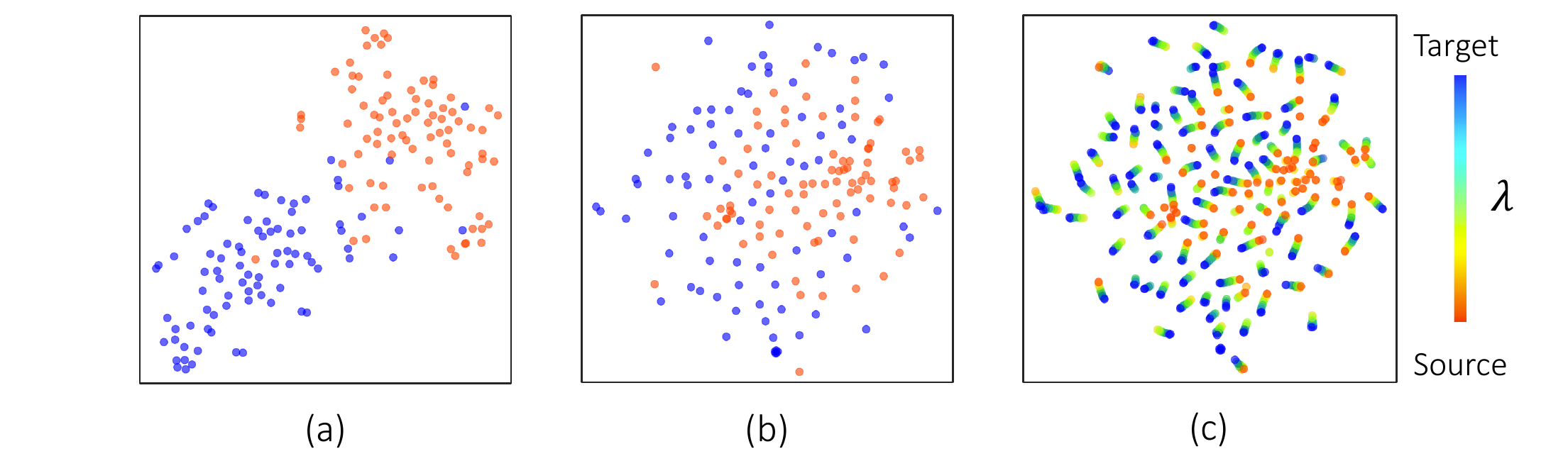}}
    \vspace{-10pt}
    \caption{\small t-SNE visualization of $S$ ({\color{red}red dots}, sampled from $\DS$) and $U$ ({\color{blue}blue dots}, sampled from $\DU$): (a) before and (b) after including \mixup in calculating the projection; (c) t-SNE of $S$, $U$, and $\mixup(S, U)$. We see a clear data transition along $\lambda$.}
    \label{fig:tsne}
    \vspace{-0.1in}
\end{figure}

We perform \mixup between labeled and pseudo-labeled data: \ie, 
between samples in $U^{({\color{red}f})}$ and $T$, and between samples in $U^{({\color{blue}g})}$ and $S$ to obtain two sets of virtual examples $\tilde{U}^{(f)}$ and $\tilde{U}^{(g)}$. We then update $\vw_{\color{red}f}$ and $\vw_{\color{blue}g}$ by SGD, 
\begin{align}
\label{eq_mist_loss}
    \vw_{\color{blue}g} \leftarrow \vw_{\color{blue}g} - \eta&\left(\nabla\mathcal{L}(\vw_{\color{blue}g}, S) +  \nabla\mathcal{L}(\vw_{\color{blue}g}, \tilde{U}^{({\color{blue}g})})\right), \\
    \vw_{\color{red}f} \leftarrow \vw_{\color{red}f} - \eta&\left(\nabla\mathcal{L}(\vw_{\color{red}f}, T) + \nabla\mathcal{L}(\vw_{\color{red}f}, \tilde{U}^{({\color{red}f})})\right), \nonumber
\end{align}
where $\eta$ is the learning rate and $\mathcal{L}$ is the averaged loss over examples. We use the cross-entropy loss.

In our experiments, we have found that \mixup can
\begin{itemize} [leftmargin=*, itemsep=1pt,topsep=2pt]
    \item effectively \textbf{denoise} an incorrect pseudo-label by mixing it with a correct one (from $S$ or $T$). The resulting $\tilde{y}$ at least contains a $\lambda$ portion of correct labels;
    \item smoothly \textbf{bridge} the domain gap between $U$ and $S$. This is done by interpolating between $U^{({\color{blue}g})}$ and $S$. The resulting $\tilde{x}$ can be seen as an intermediate example between domains.
\end{itemize}
In other words, \mixup encourages the models to behave linearly between accurately labeled and pseudo-labeled data, which reduces the undesirable oscillations caused by noisy pseudo-labels and stabilizes the predictions across domains. We note that, our usage of \mixup is fundamentally different from \cite{tang2020adversarial,xu2020adversarial,mao2019virtual,yan2020improve} that employed \mixup as auxiliary losses to augment existing DA algorithms like~\cite{ganin2016domain}.

We illustrate this in ~\cref{fig:tsne}. A model pre-trained on $D_S$ is used to generate feature embeddings. We then employ t-SNE~\cite{maaten2008visualizing} to perform two tasks simultaneously, namely clustering the embedded samples as well as projecting them into a 2D space for visualization. In (a), only {\color{red}{$S$}} sampled from $\DS$ and {\color{blue}{$U$}} sampled from $\DU$ are embedded, while in (b) and (c), additional samples from \mixup of $S$ and $U$ were added to the fold to influence t-SNE's clustering step. (b) shows only the finally projected $S$ and $U$ samples afterwards while (c) shows the additional projected \mixup samples as a function of $\lambda$. One can easily see that \mixup effectively closes the gap between the source and target domain. We summarize our proposed algorithm in \cref{alg_code}.

\begin{figure*}[t]
	\centering
    \includegraphics[width=1.0\linewidth]{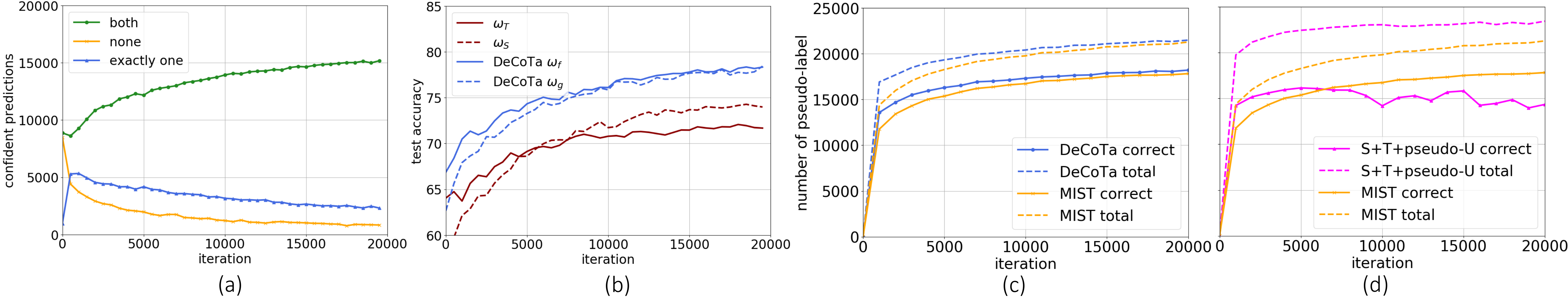}
    \vskip-10pt
    \caption{\small \textbf{Analysis on the two-task decomposition.} We use DomainNet~\cite{peng2019moment} (Real to Clipart; three-shot). (a) We show the number of test examples that \emph{both}, \emph{exactly one}, and \emph{none} of the models have high confidence on (in total, $18,325$). The two tasks hold unique expertise (\ie, there is a $14\%$ portion of the data that exactly one view is confident on), satisfying the condition of co-training in~\cref{eq_expand}. (b) We show the power of co-training: the same tasks without co-training perform worse, 
    indicating that the models benefit from each other. See~\cref{ss_muitlview} for details. The analysis is on DomainNet (R to C; three-shot) and we will clarify it. We further analyze pseudo-labels in (c) and (d). For every 1K iterations (\ie, 24K unlabeled data with possible repetition), we accumulate the number of data that have confident ($>0.5$) and correct predictions by at least one classifier. See~\cref{s_exp} for details. (c) Comparison of pseudo-label quantity and quality using \code vs. \mist. (d) \mist vs. self-training (S+T+pseudo-U). It can be observed that \code has the largest number of correct pseudo-labels.}
    \label{fig:co_training_check}
    \vspace{-0.15in}
\end{figure*}

\subsection{Constraints for Effective Co-training}
\label{ss_muitlview}
In \code, we perform co-training via a decomposition of tasks on single-view data. To explain further why \code works, we provide analysis in this subsection on the difference made by splitting the SSDA problem into two tasks for co-training. 
That is, we would like to verify that the decomposition leads to two tasks that fit into the assumption of co-training~\cite{balcan2005co}.
To begin with, we train two models: one model, $\vw_S$, is trained with $S$ and $\tilde{U}^{(S)}$ while the other model, $\vw_T$, is trained with $T$ and $\tilde{U}^{(T)}$. $\tilde{U}^{(S)}$ is obtained from applying $\vw_S$ to $U$ for pseudo-labels, follow by \mixup with $S$. The same definition goes for $\tilde{U}^{(T)}$. Essentially, both the UDA and SSL task prepare their own pseudo-labels \emph{independently} using their respective model in a procedure that is similar to self-training~\cite{lee2013pseudo,mcclosky2006effective,mcclosky2006reranking,banko2001scaling}.

\begin{table*}[h]
\small
\centering
\renewcommand{\tabcolsep}{6pt}
\renewcommand{\arraystretch}{1.12}
\caption{\small Comparing with deep co-training methods~\cite{Qiao2018DeepCF} for SSDA on DomainNet, 3-shot. (See~\cref{ss_setup} for details.)}\vspace{-0.1in}
\resizebox{0.75\linewidth}{!}{
\begin{tabular}{@{}l|c c c c c c c|c@{}}
\toprule
Method & R to C & R to P & P to C & C to S & S to P & R to S & P to R & Mean \\
\midrule
Deep Co-Training~\cite{Qiao2018DeepCF} w/o \mixup & 73.7 & 67.6 & 73.2 & 63.9 & 66.7 & 64.1 & 79.3 & 69.7\\
Deep Co-Training~\cite{Qiao2018DeepCF} with \mixup & 74.2 & 69.1 & 72.3 & 64.1 & 67.9 & 65.1 & 79.4 & 70.3\\
\vcode & 80.4 & 75.2 & 78.7 & 68.6 & 72.7 & 71.9 & 81.5 & 75.6\\
\bottomrule
\end{tabular}
}
\label{tab:compare_co_training}
\vspace{-0.15in}
\end{table*}

After training, we apply $\vw_T$ to the entire $\DU$ and compute for each $u\in\DU$ the binary confidence indicator
\begin{align}
    h_T(u) = \begin{cases}
    1 & \text{if  }  \max_c p(c|u; \vw_T) > \tau, \\ \label{eq_conf}
    0 & \text{otherwise}.
    \end{cases} 
\end{align}
Here, high confident examples will get a value $1$, otherwise $0$. 
We also apply $\vw_S$ to $\DU$ to obtain $h_S(u)$. Denote by 
$\bar{h}_T(u) = 1-{h}_T(u)$ the not function of ${h}_T(u)$, we compute the following three indicators to summarize the entire $\DU$
\begin{align}
h_\text{both}: & \sum_{u\in\DU} h_T(u)h_S(u), \nonumber\\
h_\text{one}: & \sum_{u\in\DU} h_T(u)\bar{h}_S(u) + \bar{h}_T(u)h_S(u), \label{eq_agg}\\
h_\text{none}: & \sum_{u\in\DU} \bar{h}_T(u)\bar{h}_S(u), \nonumber
\end{align}
corresponding to the number of examples that \emph{both}, \emph{exactly one}, and \emph{none} of the models have high confidence on, respectively. Intuitively, if the two models are exactly the same,  $h_\text{one}$ will be 0, meaning that they are either both confident or not on an example. On the contrary, if the two models are well optimized but hold their specialties, both $h_\text{one}$ and $h_\text{both}$ will be of high values and $h_\text{none}$ will be low.

We ran the study on DomainNet~\cite{peng2019moment}, in which we use \emph{Real} as source and \emph{Clipart} as target. (See \cref{s_exp} for details.) We consider a $126$-class classification problem, in which $|\DS|=70,358$, $|\DU|=18,325$, and $|\DT|=378$ (\ie, a three-shot setting where each class in the target domain is given three labeled samples). We initialize $\vw_S$ and $\vw_T$ with a ResNet \cite{he2016deep} pre-trained on $\DS$, and evaluate \cref{eq_conf} and \cref{eq_agg} every $500$ iterations (with a $\tau=0.5$ confidence threshold in selecting pseudo-labels.). 

\cref{fig:co_training_check} (a) shows the results. 
\emph{The two models do hold their specialties (\ie, yield different high-confident predictions)}. Even at the end of training, there is a 14\% portion of data that one model is confident on but not the other (the blue curve). Thus, if we can properly fuse their specialties during training --- one model provides the pseudo-labels to the data on which the other model is uncertain --- we are likely to jointly learn stronger models at the end.

This is indeed the core idea of our co-training proposal. Theoretically, the two ``views'' (or, tasks in our case) must satisfy certain conditions, \eg, $\epsilon$-expandability~\cite{balcan2005co}. \cite{chen2011automatic,chen2011co} relaxed it and only needed the expanding condition to hold on average in the unlabeled set, which can be formulated as follows, using $h_\text{both}$, $h_\text{one}$, and $h_\text{none}$
\begin{align}
h_\text{one}\geq \epsilon\min(h_\text{both}, h_\text{none}). \label{eq_expand}
\end{align}
To satisfy Eq.~\ref{eq_expand}, there must be sufficient examples that exactly one model is confident on so that the two models can benefit from teaching each other. Referring to~\cref{fig:co_training_check} (a) again, our two tasks consistently hold a $\epsilon$ around $2$ after the first $500$ iterations (\ie, after the models start to learn the task-specific idiosyncrasies), suggesting the feasibility of applying co-training to our decomposition. The power of co-training is clearly illustrated in~\cref{fig:co_training_check} (b). The two models without co-training, \underline{$\vw_{T}$ and $\vw_{S}$}, perform worse than their co-training counterparts,  \underline{$\vw_{\color{red}f}$ and $\vw_{\color{blue}g}$} (see~\cref{ss_code}, \cref{eq_code}, \cref{eq_mist_loss}), even using the same architecture and data.

\subsection{Comparing to Other Co-training Approaches}
\label{CODA}
With our approach outlined, it is worthwhile to contrast \code with prior co-training work in domain adaptation. In particular, \code is notably different from the approach known as Co-training for DA (CODA) \cite{chen2011co}. While CODA also utilizes co-training for SSDA using single-view data, it differs from \code fundamentally as follow: 
\begin{enumerate} [leftmargin=*, itemsep=1pt,topsep=2pt]
    \item CODA takes a feature-centric view in that the two \emph{artificial} views in its co-training procedure are constructed by decomposing the feature dimensions into two mutually exclusive subsets. \code on the other hand achieves effective co-training with a two-task decomposition.
    \item The two views in CODA do not exchange high confident pseudo-labels in a mini-batch fashion like \code. Nor does CODA utilize \mixup, which we have shown to be valuable for SSDA. Instead, CODA explicitly conducts feature alignment by minimizing the difference between the distributions of the source and target domains.
    \item CODA trains a logistic regression classifier. In the era of deep learning, while co-training has been used in multiple vision tasks, \code is \emph{the first work in SSDA} utilizing deep learning, co-training, and mixup in a cohesive and principled fashion, achieving state of the art performance.
\end{enumerate}
Since CODA is not deep learning based, to further justify the efficacy of \code, we took the deep co-training work described in \cite{Qiao2018DeepCF} that was designed for semi-supervised image recognition, and customize it for SSDA. \cite{Qiao2018DeepCF} constructs multi-views for co-training via two different adversarial perturbations on the same image samples, after which the two networks are trained to make different mistakes on the same adversarial examples. For fair comparison, we compare~\cite{Qiao2018DeepCF} both with and without \mixup, using the DomainNet~\cite{peng2019moment} dataset.
The results are given in \cref{tab:compare_co_training}.
\code outperforms~\cite{Qiao2018DeepCF} by a margin. See~\cref{ss_setup} for detailed setups.
\section{Experiments}
\label{s_exp}

\begin{table*}[]
    \footnotesize
    \centering
    \renewcommand{\tabcolsep}{6pt}
    \renewcommand{\arraystretch}{1.2}
     \caption{\small Accuracy on DomainNet (\%) for three-shot setting with $4$ domains, using ResNet-34.
    }\vspace{-0.1in}
    \resizebox{0.6\linewidth}{!}{
    \begin{tabular}{@{}l|c c c c c c c|c@{}}
    \toprule
        Method & R to C & R to P & P to C & C to S & S to P & R to S & P to R & Mean \\
        \midrule
         S+T & 60.8 & 63.6 & 60.8 & 55.6 & 59.5 & 53.3 & 74.5 & 61.2\\
         DANN~\cite{ganin2016domain} & 62.3 & 63.0 & 59.1 & 55.1 & 59.7 & 57.4 & 67.0 & 60.5 \\
         ENT~\cite{saito2019semi} & 67.8 & 67.4 & 62.9 & 50.5 & 61.2 & 58.3 & 79.3 & 63.9\\
         MME~\cite{saito2019semi} & 72.1 & 69.2 & 69.7 & 59.0 & 64.7 & 62.2 & 79.0 & 68.0\\
         UODA~\cite{qin2020opposite} & 75.4 & 71.5 & 73.2 & 64.1 & 69.4 & 64.2 & 80.8 & 71.2\\
         APE~\cite{kim2020attract} & 76.6 & 72.1 & 76.7 & 63.1 & 66.1 & 67.8 & 79.4 & 71.7 \\
         ELP~\cite{huang2020effective} & 74.9 &  72.1 & 74.4 & 64.3 & 69.7 & 64.9 & 81.0 & 71.6\\
         \vcode & \textbf{80.4} & \textbf{75.2} & \textbf{78.7} & \textbf{68.6} & \textbf{72.7} & \textbf{71.9} & \textbf{81.5} & \textbf{75.6}\\
         \bottomrule
    \end{tabular}
    }
   
    \label{tab:sota_domainet}
    \vspace{-0.1in}
\end{table*}

\begin{table*}[]
    \footnotesize
    \centering
    \renewcommand{\tabcolsep}{6pt}
    \renewcommand{\arraystretch}{1.2}
        \caption{\small Accuracy on Office-Home (\%) for three-shot setting with $4$ domains, using VGG-16.}\vspace{-0.1in}
\resizebox{0.92\textwidth}{!}{%
\begin{tabular}{@{}l|c c c c c c c c c c c c|c@{}}
    \toprule
    Method & R to C & R to P & R to A & P to R & P to C & P to A & A to P &
     A to C & A to R & C to R & C to A & C to P & Mean \\
     \midrule
     S+T & 49.6 &  78.6 &  63.6 & 72.7 & 47.2 & 55.9 & 69.4 & 47.5 &  73.4 & 69.7 & 56.2 &  70.4 & 62.9\\
     DANN~\cite{ganin2016domain} & 56.1 & 77.9 & 63.7 & 73.6 & 52.4 & 56.3 & 69.5 & 50.0 &  72.3 & 68.7 & 56.4 & 69.8 &  63.9\\
     ENT~\cite{saito2019semi} & 48.3 &  81.6 & 65.5 & 76.6 & 46.8 & 56.9 & 73.0 &  44.8 & 75.3 & 72.9 & 59.1 & 77.0 & 64.8\\
     MME~\cite{saito2019semi} & 56.9 & 82.9 & 65.7 & 76.7 & 53.6 &  59.2 & 75.7 &  54.9 & 75.3 & 72.9 & 61.1 & 76.3 & 67.6\\
     UODA~\cite{qin2020opposite} & 57.6 & 83.6 & 67.5 & \textbf{77.7} & 54.9 & \textbf{61.0} & 77.7 & \textbf{55.4} & \textbf{76.7} & 73.8 &  61.9 & \textbf{78.4} & 68.9\\
     APE~\cite{kim2020attract} & 56.0 & 81.0 & 65.2 & 73.7 & 51.4 & 59.3 & 75.0 & 54.4 & 73.7 & 71.4 & 61.7 & 75.1 & 66.5 \\
     ELP~\cite{huang2020effective} & 57.1 & 83.2 & 67.0 & 76.3 & 53.9 & 59.3 & 75.9 & 55.1 & 76.3 & 73.3 & 61.9 & 76.1 & 68.0 \\
     \vcode & \textbf{59.9} & \textbf{83.9} & \textbf{67.7} & 77.3 & \textbf{57.7} & 60.7 & \textbf{78.0} & 54.9 & 76.0 & \textbf{74.3} & \textbf{63.2} & \textbf{78.4} & \textbf{69.3}\\
     \bottomrule
    \end{tabular}
    \label{tab:sota_office_hom}
}
\vspace{-0.12in}
\end{table*}

\label{ss_setup}

We consider the one-/three-shot settings, following~\cite{saito2019semi}, where each class is given one or three labeled target examples. We train with $\DS$, $\DT$, and unlabeled $\DU$. We then reveal the true label of $\DU$ for evaluation.

\noindent\textbf{Datasets.}
We use \textbf{DomainNet} \cite{peng2019moment}, a large-scale benchmark dataset for domain adaptation that has $345$ classes and $6$ domains. We follow~\cite{saito2019semi}, using a $126$-class subset with $4$ domains (\ie, R: Real, C: Clipart, P: Painting, S: Sketch.) and report $7$ different adaptation scenarios. 
We also use \textbf{Office-Home}~\cite{venkateswara2017deep}, another benchmark that contains $65$ classes, with $12$ adaptation scenarios constructed from $4$ domains (\ie, R: Real world, C: Clipar t, A: Art, P: Product).

\noindent\textbf{Implementation details.}
We implement using Pytorch \cite{paszke2017automatic}. 
We follow~\cite{saito2019semi} to use ResNet-34 \cite{he2016deep} on DomainNet and VGG-16~\cite{simonyan2014very} on Office-Home. We also provide ResNet-34 results on Office-Home in order to fairly compare with ~\cite{kim2020attract} in supplementary. The networks are pre-trained on ImageNet~\cite{deng2009imagenet,russakovsky2015imagenet}.
We follow~\cite{saito2019semi,ranjan2017l2} to 
replace the last linear layer with a $K$-way cosine classifier (\eg, $K=126$ for DomainNet) and train it at a fixed temperature ($0.05$ in all our experiments). We initialize $\vw_f$ with a model first fine-tuned on $\DS$, and initialize $\vw_g$ with a model first fine-tuned on $\DS$ and then fine-tuned on $\DT$. We do so to encourage the two models to be different at the beginning.
At each iteration, we sample three mini-batches $S\subset\DS$, $T\subset\DT$, and $U\subset\DU$ of equal sizes $B=24$ (cf. \cref{ss_mixup}).
We set the confidence threshold $\tau=0.5$, and beta distribution coefficient $\alpha=1.0$. We use SGD with momentum of $0.9$ and an initial learning rate of $0.001$, following~\cite{saito2019semi}. We train for 50K/10K iterations on DomainNet/Office-Home.
We note that, \code does not increase the training time since at each iteration, it only updates and learns from the pseudo-labels of the current mini-batch of unlabeled data, not the entire unlabeled data.

\begin{table*}[h]
\small
\centering
    \renewcommand{\tabcolsep}{4pt}
    \renewcommand{\arraystretch}{1.1}
    \caption{\small Ablation Study (three shots). (a)-(b): comparison of \mist and \code and the vanilla ensemble of two independently trained \mist; (c): comparison of Two-view \mist (without co-training) and \code; (d) comparison of \mist and S+T+pseudo-U without \mixup; (e) each model of \code on the source domain test data, comparing to supervised training on source (S), average of DomainNet. All accuracy in ($\%$).}
    \vspace{-0.1in}
    \begin{subtable}{1.0\textwidth}
    \centering
    \caption{\small Comparing \mist, Vanilla-Ensemble of two \mist (with different initialization), and \code on DomainNet}
    \resizebox{0.56\linewidth}{!}{
        \begin{tabular}{@{}l|c c c c c c c|c@{}}
            \toprule
            Method & R to C & R to P & P to C & C to S & S to P & R to S & P to R & Mean \\
            \midrule
            \vmist & 78.1 & 75.2 & 76.7 & 68.3 & 72.6 & 71.5 & 79.8& 74.6\\
            Vanilla-Ensemble & 79.7 & 75.0 & 77.2 & 68.4 & 72.1 & 70.8 & 79.7 & 74.7\\
            \vcode & 80.4 & 75.2 & 78.7 & 68.6 & 72.7 & 71.9 & 81.5 & 75.6\\
            \bottomrule
        \end{tabular}
        }
        \vspace{0.08in}
    \end{subtable}
    
    \begin{subtable}{1.0\textwidth}
    \centering
    \renewcommand{\tabcolsep}{4pt}
    \renewcommand{\arraystretch}{1.1}
    \caption{\small Comparing \mist, Vanilla-Ensemble of two \mist (with different initialization), and \code on Office-Home}
    \resizebox{0.85\linewidth}{!}{
    \begin{tabular}{@{}l|c c c c c c c c c c c c|c@{}}
    \toprule
    Method & R to C & R to P & R to A & P to R & P to C & P to A & A to P & A to C & A to R & C to R & C to A & C to P & Mean \\
    \midrule
    \vmist & 54.7 & 81.2 & 64.0 & 69.4 & 51.7 & 58.8 & 69.1 & 47.6 & 70.6 & 65.3 & 60.8 & 73.8 & 63.9\\
    Vanilla-Ensemble & 56.1 & 81.8 & 63.4 & 72.9 & 54.1 & 55.1 & 74.2 & 49.5 & 72.1 & 67.4 & 55.2 & 75.6 & 64.7\\
    \vcode & 59.9 & 83.9 & 67.7 & 77.3 & 57.7 & 60.7 & 78.0 & 54.9 & 76.0 & 74.3 & 63.2 & 78.4 & 69.3\\
    \bottomrule
     \end{tabular}
     }
     \vspace{0.08in}
     \end{subtable}
     
    \begin{subtable}{1.0\textwidth}
     \centering
     \renewcommand{\tabcolsep}{4pt}
    \renewcommand{\arraystretch}{1.1}
    \caption{\small Comparing the decomposed tasks trained independently to using \code}\vspace{-0.05in}
    \resizebox{0.66\linewidth}{!}{
         \begin{tabular}{@{}l|c|c c c c c c c|c@{}}
            \toprule
            Method & Task & R to C & R to P & P to C & C to S & S to P & R to S & P to R & Mean \\
            \midrule
            \multirow{3}{*}{\makecell[l]{Decomposed tasks\\(without co-training)}} 
            & $\vct{w}_f$ & 72.1 & 65.7 & 71.8 & 61.0 & 63.0 & 59.9 & 75.9 & 67.0\\
            &$\vct{w}_g$ & 76.3 & 72.2 & 70.3 & 63.7 & 69.4 & 66.9 & 76.1 & 70.7\\
            & Ensemble & 77.3 & 72.0 & 75.1 & 65.7 & 69.3 & 66.1 & 78.7 & 72.0\\
            \midrule
            \multirow{3}{*}{\vcode} & $\vct{w}_f$ & 80.1 & 74.6 & 78.6 & 68.4 & 72.5 & 71.2 & 81.1 & 75.2 \\
            & $\vct{w}_g$ & 80.0 & 74.5 & 78.4 & 68.3 & 72.2 & 71.3 & 80.6 & 75.0\\
            & Ensemble & 80.4 & 75.2 & 78.7 & 68.6 & 72.7 & 71.9 & 81.5 & 75.6\\
             \bottomrule
        \end{tabular}
        }
    \vspace{0.08in}
    \end{subtable}

    \begin{subtable}{0.95\textwidth}
    \centering
    \begin{minipage}{.58\linewidth}
    \centering
    \caption{Comparing \mist and the S+T+pseudo-U with no \mixup on DomainNet}\vspace{-0.05in}
    \resizebox{1.0\linewidth}{!}{
        \begin{tabular}{@{}l|c c c c c c c|c@{}}
            \toprule
            Method & R to C & R to P & P to C & C to S & S to P & R to S & P to R & Mean \\
            \midrule
             \textbf{S+T+pseudo-U} & 70.0 & 67.2 & 68.3 & 57.2 & 61.1 & 58.7 & 71.2 & 65.6 \\
             \vmist & 78.1 & 75.2 & 76.7 & 68.3 & 72.6 & 71.5 & 79.8& 74.6\\
             \bottomrule
        \end{tabular}
        }
        
    \end{minipage}
    \hfill
    \begin{minipage}{.34\linewidth}
    \centering
    \caption{Accuracy on source domain}\vspace{-0.05in}
    \resizebox{0.65\linewidth}{!}{
        \begin{tabular}{@{}ccc|c@{}}
        \toprule
        $\vw_f$ & $\vw_g$ & \vcode & \textbf{S} \\
        \midrule
        65.3 & 98.2 & 93.5 & 98.8 \\
        \bottomrule
        \\
        \end{tabular}
        }
        
    \end{minipage}

    \end{subtable}
\label{ablation_study}
\vspace{-0.1in}
\end{table*}

\noindent\textbf{Baselines.}
We compare to four state-of-the-art SSDA approaches, \textbf{MME}~\cite{saito2019semi},  \textbf{UODA}~\cite{qin2020opposite}, \textbf{APE}~\cite{kim2020attract}, and
\textbf{ELP}~\cite{huang2020effective}.
We also compare to \textbf{S+T}, a model trained with $\DS$ and $\DT$, without using $\DU$. Additionally, we compare to \textbf{DANN} \cite{ganin2016domain} (domain adversarial learning) and \textbf{ENT}~\cite{grandvalet2005semi} (entropy minimization), both of which are important prior work on UDA. We modify them such that $\DS$ and $\DT$ are used jointly to train the classifier, following~\cite{saito2019semi}. We denote by \textbf{S} the model trained only with the source data $\DS$.

\noindent\textbf{Variants of our approach.}
We consider variants of our approach for extensive ablation studies. We first introduce a model we called \mixup Self-Training (\mist). \mist is trained as follows
\begin{align}
\vw \leftarrow \vw - \eta&\nabla\mathcal{L}(\vw, S) + \nabla\mathcal{L}(\vw, T) \\
    & + \nabla\mathcal{L}(\vw, \tilde{U}_S^{(\vw)}) + \nabla\mathcal{L}(\vw, \tilde{U}_T^{(\vw)})), \nonumber
\label{eq:mist}
\end{align}
where $\tilde{U}_S^{(\vw)}$ and $\tilde{U}_T^{(\vw)}$ are pseudo-labels obtained from $\vw$, followed by \mixup with $S$ and $T$, respectively. \mist basically lumps all the pseudo and hard labeled samples together during training, and is intended for comparing with the effect of co-training. \textbf{S+T+pseudo-U} is the model trained with self-training, but without \mixup.
\textbf{Two-view \mist} is the direct ensemble of independently trained models, one for each view, using \mist (cf. \cref{ss_muitlview}). 
\textbf{Vanilla-Ensemble} is the ensemble model by combining two $\mist$ trained on $\DS$, $\DT$, and $\DU$ but with different initialization. 
For all the variants that train only one model, 
we initialize it with a pre-trained model fine-tuned on $\DS$ and then fine-tuned on $\DT$. Otherwise, we initialize the two models in the same way as \code. \emph{We note that, for any methods that involve two models, we perform ensemble on their output probability.}

\noindent\textbf{Main results.} 
We summarize the comparison with baselines in \cref{tab:sota_domainet} and \cref{tab:sota_office_hom}. 
We mainly report the three-shot results and leave the one-shot results in the supplementary material.
\code outperforms other methods by a large margin on DomainNet, and outperforms all methods on Office-Home (mean). The smaller gain on Office-Home may be due to its smaller data size and limited scenes. DomainNet is larger and more diverse; the significant improvement on it is a stronger indicator of the effectiveness of our algorithm.

We further provide detailed analysis on \code. We mainly report the DomainNet three-shot results. Other detailed results can be found in the supplementary material.

\noindent\textbf{Task decomposition.}
We first compare \code to \mist. As shown in ~\cref{ablation_study} (a)-(b), \code outperforms \mist by $1\%$ on DomainNet and $5\%$ on Office-Home on the three-shot setup. \cref{fig:co_training_check} (c) further shows the number of pseudo-labels  involved in model training (those with confidence larger than $\tau=0.5$). We see that \code always generates more pseudo-label data with a higher accuracy than \mist (also in~\cref{fig:co_training_check} (b)), justifying our claim that the decomposition helps keep $\DS$'s and $\DT$'s specialties, producing high confident predictions on more unlabeled data as a result.

\noindent\textbf{Co-training.}
We compare \code to \textbf{two-view \mist}. Both methods decompose the data into a SSL and a UDA task. The difference is in  how the pseudo-label set was generated (cf. \cref{eq_code}): \textbf{Two-view \mist} constructs each set independently (cf. \cref{ss_muitlview}). \code outperforms {two-view \mist} by a margin, not only on ensemble, but also on each view alone, justifying the effectiveness of two models exchanging their specialties to benefit each other. As in \cref{ablation_study} (c), each model of \code outperforms \mist.

\noindent\textbf{\mixup.} We examine the importance of \mixup. Specifically, we compare \mist and \textbf{S+T+pseudo-U}. The second model trains in the same way as \mist, except that it does not apply \mixup. On DomainNet (3-shot), \mist outperforms \textbf{S+T+pseudo-U} by \textbf{9\%} on average.
We attribute this difference to the \emph{denoising} effect by \mixup: \mixup is performed after the pseudo-label set is defined, so it does not directly affect the number of pseudo-labels, but the quality. We further calculate the number of correctly assigned pseudo-labels along training, as shown in~\cref{fig:co_training_check} (d). With \mixup, the correct pseudo-label pool boosts consistently. In contrast, \textbf{S+T+pseudo-U} reinforces itself with wrongly assigned pseudo-labels; the percentage thus remains constantly low. Comparison results are shown in~\cref{ablation_study} (d).

\noindent\textbf{Comparison to vanilla model ensemble.}
Since \code combines $\vw_f$ and $\vw_g$ in making predictions, for a fair comparison  we train two \mist models (both use $\DS+\DT+\DU$), each with different initialization, and perform model ensemble. As shown in~\cref{ablation_study}~(a)-(b), \code outperforms this vanilla model ensemble, especially on Office-Home, suggesting that our improvement does not simply come from model ensemble, but from co-training.

\noindent\textbf{On the ``two-classifier-convergence'' problem \cite{yu2019does}.}
\code is based on co-training and thus does not suffer the problem.
This is shown in \autoref{ablation_study} (a, b): \mist and Vanilla-Ensemble are based on self-training and \code outperformed them. Even at the end of training when two classifiers have similar accuracy (see \autoref{ablation_study} (c)), combining them still boosts the accuracy: \ie, they make different predictions.

\noindent\textbf{Results on the source domain.} While $\vw_f$ and $\vw_g$ have similar accuracy on $\DU$, the fact that $\vw_f$ does not learn from $\DS$ suggest their difference in classifying source domain data. We verify this in~\cref{ablation_study}~(e), where we apply each model individually on a hold-out set from the source domain (provided by DomainNet). We see that $\vw_g$ clearly dominates $\vw_f$. Its accuracy is even on a par with a model trained only on $\DS$, showing one advantage of \code  --- the model can keep its discriminative ability on the source domain.
\section{Conclusion}
\label{s_disc}
\vspace{-0.08in}

We introduce \vcode, a simple yet effective approach for semi-supervised domain adaptation (SSDA). Our key contribution is the novel insight that the two sources of supervisions (\ie, the labeled target and labeled source data) are inherent different and should not be combined directly.
\code thus explicitly decomposes SSDA into two tasks (\ie, views), 
a semi-supervised learning task and an unsupervised domain adaptation task, in which each supervision can be better leveraged.
To encourage knowledge sharing and integration between the two tasks, we employ co-training, a well-established technique that allows for distinct views to learn from each other. We provided empirical evidence that the two tasks satisfy the theoretical condition of co-training, which makes \code well founded, simple (without adversarial learning), and superior in performance.

\smallskip
\noindent\textbf{Acknowledgement.} This research was supported by independent awards from Facebook AI, NSF (III-1618134, III- 1526012, IIS-1149882, IIS-1724282, TRIPODS-1740822, OAC-1934714, DMR-1719875), ONR DOD (N00014-17-1-2175), DARPA SAIL-ON (W911NF2020009), and the Bill and Melinda Gates Foundation. We are thankful for generous support by Ohio Supercomputer Center and AWS Cloud Credits for Research.

{\small
\bibliographystyle{ieee_fullname}
\bibliography{egbib}
}

\clearpage
\appendix
\begin{strip}
\centering
	\textbf{\Large Deep Co-Training with Task Decomposition\\for Semi-Supervised Domain Adaptation \\[0.2em] {\large (Supplementary Material)}}
	\vskip 10pt
	\author{
$\textbf{Luyu Yang}^{1}, \quad \textbf{Yan Wang}^{2}, \quad\textbf{Mingfei Gao}^{3}$,\\[0.2em]
$\textbf{Abhinav Shrivastava}^{1}, \textbf{Kilian Q. Weinberger}^{2}$, $\textbf{Wei-Lun Chao}^{4}, \textbf{Ser-Nam Lim}^{5}$\\[0.5em]
\medskip
$^{1}\text{University of Maryland}$\quad
$^{2}\text{Cornell University}$\\[0.2em]
$^{3}\text{Salesforce Research}$\quad
$^{4}\text{Ohio State University}$\quad $^{5}\text{Facebook AI}$
\vspace{10pt}
}
\end{strip}

We provide details omitted in the previous sections.
\begin{itemize} [leftmargin=*,noitemsep,topsep=0pt]
    \item \cref{suppl-sec:exp_s}: additional details on experimental setups (cf. \autoref{s_exp} of the main paper).
    \item \cref{suppl-sec:exp_r}: additional details on experimental results (cf. \autoref{s_exp} of the main paper).
\end{itemize}

\section{Experimental Setups}
\label{suppl-sec:exp_s}

\begin{figure}[b]
    \vspace{-0.1in}
	\centerline{
    \includegraphics[width=0.9\linewidth]{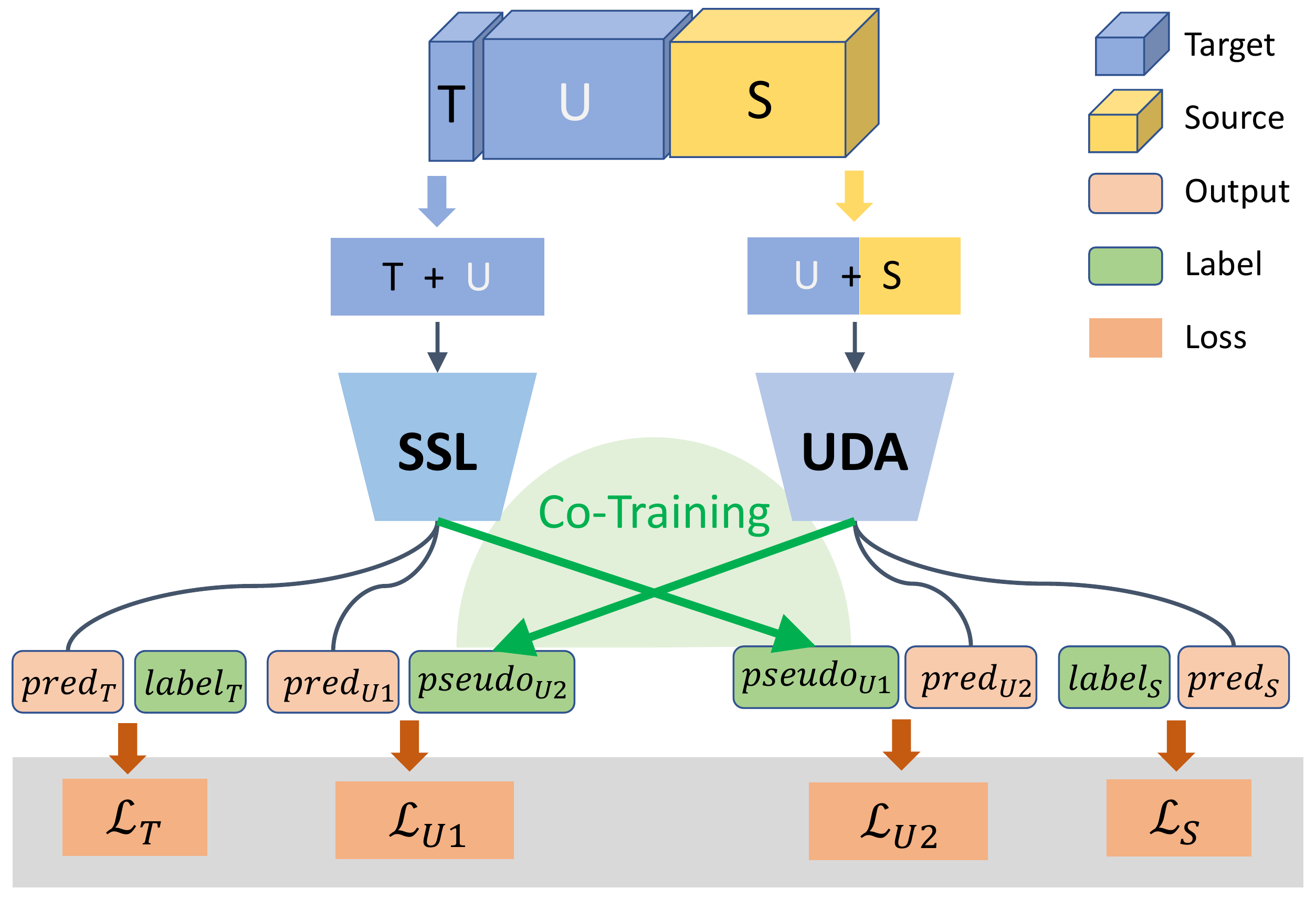}}
    \caption{\small The overall framework of \vcode. It decomposes the SSDA task into SSL and UDA tasks that exchange pseudo-labels for unlabeled target \emph{U}.}
    \label{fig:overall_framework}
\end{figure}

\begin{figure}[t]
\vspace{-0.1in}
	\centering
    \includegraphics[width=1\linewidth]{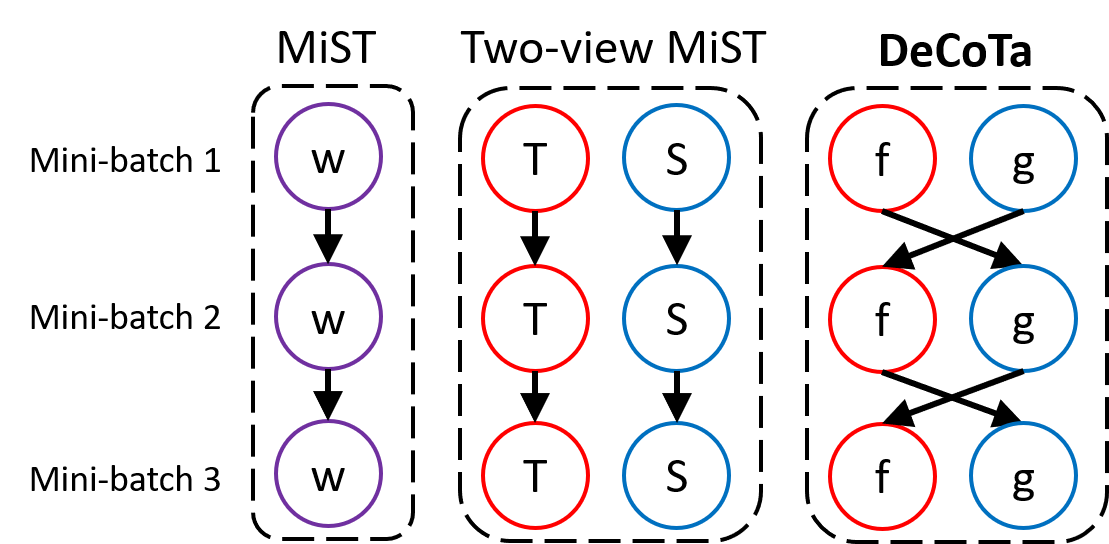}
     \caption{\small Comparison among \vmist, two-view \vmist (\ie, two-task \vmist), and \vcode. The color on the circles means the labeled data: {\color{red}red} for $\DT$, {\color{blue}blue} for $\DS$, and {\color{purple}purple} for both. The arrows indicate which model provides the pseudo-labels for which model to learn from.}
    \label{suppl-fig:method_comp}
\end{figure}

In section 5 of the main paper, we compare variants of our approach, including \vmist, two-view \vmist, and \vcode. Here we give some more discussions. These three methods are different by 1) how many classifiers they train; 2) what labeled data they use; 3) which classifier provides the pseudo-labels.
\cref{suppl-fig:method_comp} gives an illustrative comparison. \cref{fig:overall_framework} illustrates the framework pipeline of \vcode.
\begin{itemize} [leftmargin=*,noitemsep,topsep=0pt]
    \item \vmist learns a single model $\vw$, using both labeled source data $\DS$ and labeled target data $\DT$. \vmist also updates $\vw$ using pseudo-labels on the unlabeled target data $\DU$, where the pseudo-labels are predicted by the current $\vw$.
    \item \textbf{Two-view} \vmist (\ie, two-task \vmist) learns two models, $\vw_{\color{red}T}$ and $\vw_{\color{blue}S}$ (cf. subsection 3.2 of the main paper). $\vw_{\color{red}T}$ is updated using $\DT$ and pseudo-labeled data on $\DU$, where the pseudo-labels are predicted by the current $\vw_{\color{red}T}$. $\vw_{\color{blue}S}$ is updated using $\DS$ and pseudo-labeled data on $\DU$, where the pseudo-labels are predicted by the current $\vw_{\color{blue}S}$.
    \item \vcode learns two models, $\vw_{\color{red}f}$ and $\vw_{\color{blue}g}$. $\vw_{\color{red}f}$ is updated using $\DT$ and pseudo-labeled data on $\DU$, where the pseudo-labels are predicted by the current $\vw_{\color{blue}g}$. $\vw_{\color{blue}g}$ is updated using $\DS$ and pseudo-labeled data on $\DU$, where the pseudo-labels are predicted by the current $\vw_{\color{red}f}$.
\end{itemize}

\vcode has two hyper-parameters: the confidence threshold $\tau$ (cf. \autoref{eq_code} of the main paper) and $\alpha$ in \mixup (cf. \autoref{eq_mixup} of the main paper). We follow~\cite{saito2019semi} to select these hyper-parameters using three other labeled examples per class in the target domain. Specifically, we only select hyper-parameters based on DomainNet three-shot setting, Real to Clipart. We then fix the selected hyper-parameters, $\tau=0.5$ and $\alpha=1.0$, for all other experiments.

\begin{table*}[t]
    \footnotesize
    \centering
    \renewcommand{\tabcolsep}{6pt}
    \renewcommand{\arraystretch}{1.2}
    \caption{\small Accuracy on DomainNet (\%) for the one-shot setting with four domains, using ResNet-34.}\vspace{-0.1in}
    \begin{tabular}{@{}l|c c c c c c c|c@{}}
    \toprule
        Method & R to C & R to P & P to C & C to S & S to P & R to S & P to R & Mean \\
        \midrule
         S+T & 58.1 & 61.8 & 57.7 & 51.5 & 55.4 & 49.1 & 73.1 & 58.1\\
         DANN~\cite{ganin2016domain} & 61.2 & 62.3 & 56.4 & 54.0 & 57.9 & 55.9 & 65.6& 59.0\\
         ENT~\cite{saito2019semi} & 60.0 & 60.2 & 54.9 & 48.3 & 55.8 & 49.4 & 74.4 & 57.6\\
         MME~\cite{saito2019semi} & 69.5 & 68.1 & 64.4 & 56.7 & 62.0 & 59.2 & 76.9 & 65.3\\
         UODA~\cite{qin2020opposite} & 72.7 & 70.3 & 69.8 & 60.5 & 66.4 & 62.7 & 77.3 & 68.5\\
         APE~\cite{kim2020attract} & 70.4 & 70.8 & 72.9 & 56.7 & 64.5 & 63.0 & 76.6 & 67.6\\
         ELP~\cite{huang2020effective} & 72.8 &  70.8 & 72.0 & 59.6 & 66.7 & 63.3 & 77.8 & 69.0\\
         \vcode & \textbf{79.1} & \textbf{74.9} & \textbf{76.9} & \textbf{65.1} & \textbf{72.0} & \textbf{69.7} & \textbf{79.6} & \textbf{73.9}\\
         \bottomrule
        \end{tabular}
    \label{tab:sota_domainet_one_shot}
\end{table*}

\begin{table*}[t]
    \footnotesize
    \centering
    \renewcommand{\tabcolsep}{6pt}
    \renewcommand{\arraystretch}{1.2}
    \caption{\small Accuracy on Office-Home (\%) for the one-shot setting with four domains, using VGG-16.}\vspace{-0.1in}
    \begin{tabular}{@{}l|c c c c c c c c c c c c|c@{}}
    \toprule
    Method & R to C & R to P & R to A & P to R & P to C & P to A & A to P &
     A to C & A to R & C to R & C to A & C to P & Mean \\
     \midrule
     S+T & 39.5 & 75.3 & 61.2 & 71.6 & 37.0 & 52.0 & 63.6 & 37.5 & 69.5 &  64.5 & 51.4 & 65.9 & 57.4\\
     DANN~\cite{ganin2016domain} & \textbf{52.0} & 75.7 &  62.7 & 72.7 & 45.9 &  51.3 & 64.3 & 44.4 &  68.9 &  64.2 & 52.3 & 65.3 & 60.0\\
     ENT~\cite{saito2019semi} & 23.7 & 77.5 &  64.0 & 74.6 & 21.3 & 44.6 & 66.0 & 22.4 & 70.6 & 62.1 & 25.1 & 67.7 & 51.6\\
     MME~\cite{saito2019semi} & 49.1 & 78.7 &  65.1 & 74.4 & 46.2 & 56.0 &  68.6 & 45.8 & 72.2 & 68.0 & 57.5 & 71.3 & 62.7\\
     UODA~\cite{qin2020opposite} & 49.6 & 79.8 & \textbf{66.1} & 75.4 & 45.5 & \textbf{58.8} & \textbf{72.5} & 43.3 & \textbf{73.3} & 70.5 & \textbf{59.3} & 72.1 & \textbf{63.9}\\
     ELP~\cite{huang2020effective} & 49.2 & 79.7 & 65.5 & 75.3 & 46.7 & 56.3 & 69.0 & \textbf{46.1} & 72.4 & 68.2 & 67.4 & 71.6 & 63.1\\
     \vcode & 47.2 & \textbf{80.3} & 64.6 & \textbf{75.5} & \textbf{47.2} & 56.6 & 71.1 & 42.5 & 73.1 & \textbf{71.0} & 57.8 & \textbf{72.9} & 63.3\\
     \bottomrule
    \end{tabular}
    \label{tab:sota_office_hom_one_shot}
\end{table*}

\begin{table*}[t]
\footnotesize
\renewcommand{\tabcolsep}{6pt}
\renewcommand{\arraystretch}{1.2}
\centering
\caption{\small Accuracy on Office-Home (\%) for the three-shot setting with four domains, using ResNet-34.}\vspace{-0.1in}
\begin{tabular}{@{}l|c c c c c c c c c c c c|c@{}}
     \toprule
     Method & R to C & R to P & R to A & P to R & P to C & P to A & A to P &
            A to C & A to R & C to R & C to A & C to P & Mean \\
    \midrule
     S+T & 55.7 & 80.8 & 67.8 & 73.1 & 53.8 & 63.5 & 73.1 & 54.0 & 74.2 & 68.3 & 57.6 & 72.3 & 66.2\\
     DANN~\cite{ganin2016domain} & 57.3 & 75.5 & 65.2 & 69.2 & 51.8 & 56.6 & 68.3 & 54.7 & 73.8 & 67.1 & 55.1 & 67.5 & 63.5\\
     ENT~\cite{saito2019semi} & 62.6 & 85.7 & 70.2 & 79.9 & 60.5 & 63.9 & 79.5 & 61.3 & 79.1 & 76.4 & 64.7 & 79.1 & 71.9\\
     MME~\cite{saito2019semi} & 64.6 & 85.5 & 71.3 & 80.1 & 64.6 & 65.5 & 79.0 & 63.6 & 79.7 & 76.6 & 67.2 & 79.3 & 73.1\\
     APE~\cite{kim2020attract} & 66.4 & 86.2 & 73.4 & 82.0 & 65.2 & 66.1 & 81.1 & 63.9 & 80.2 & 76.8 & 66.6 & 79.9 & 74.0 \\
     \vcode & \textbf{70.4} & \textbf{87.7} & \textbf{74.0} & \textbf{82.1} & \textbf{68.0} & \textbf{69.9} & \textbf{81.8} & \textbf{64.0} & \textbf{80.5} & \textbf{79.0} & \textbf{68.0} & \textbf{83.2} & \textbf{75.7}\\
    \bottomrule
    \end{tabular}
     \label{tab:resnet_office_hom}
\end{table*}

%
\vspace{-0.1in}
\begin{table*}[t]
    \footnotesize
    \centering
    \renewcommand{\tabcolsep}{6pt}
    \renewcommand{\arraystretch}{1.2}
    \caption{\small SSDA results on Office-31, on two scenarios (following~\cite{saito2019semi}).}\vspace{-0.1in}
    \resizebox{0.5\linewidth}{!}{
    \begin{tabular}{@{}l c c c c@{}}
    \toprule
    \multirow{2}{*}{Method} & \multicolumn{2}{c}{Webcam (W) to Amazon (A)} & \multicolumn{2}{c@{}}{DSLR
(D) to Amazon (A)} \\
    \cmidrule(lr){2-3}
    \cmidrule(l){4-5}
    {} & 1-shot & 3-shot & 1-shot & 3-shot\\
    \midrule
     S+T & 69.2 & 73.2 & 68.2 & 73.3\\
     DANN~\cite{ganin2016domain} & 69.3 & 75.4 & 70.4 & 74.6\\
     ENT~\cite{saito2019semi} & 69.1 & 75.4 & 72.1 & 75.1\\
     MME~\cite{saito2019semi} & 73.1 & 76.3 & 73.6 & 77.6\\
     Ours & \textbf{76.0} & \textbf{76.8} & \textbf{74.2} & \textbf{78.3}\\
    \bottomrule
    \end{tabular}
    }
    \label{tab:office_result}
    \vspace{-0.1in}
\end{table*}

\smallskip
\section{Experimental Results}
\label{suppl-sec:exp_r}

\subsection{Main results on the one-shot setting}
We report the comparison with baselines in the one-shot setting on DomainNet in \cref{tab:sota_domainet_one_shot} and Office-Home in \cref{tab:sota_office_hom_one_shot}. \vcode outperforms the state-of-the-art methods by $4.9\%$ on DomainNet (ResNet-34), while performs slightly worse than~\cite{qin2020opposite} by $0.6\%$ on Office-Home (VGG-16). Nevertheless, \vcode attains the highest accuracy on $5$ adaptation scenarios of Office-Home in the one-shot setting.

\subsection{Office-Home results on other backbones}
We report the comparison with baselines on Office-Home using a ResNet-34 backbone in \cref{tab:resnet_office_hom}, following \cite{kim2020attract}\footnote{Most existing papers only reported Office-Home results using VGG-16. We followed \cite{kim2020attract} to further report ResNet-34. Some algorithms reported in \cref{tab:sota_office_hom} are missing in \cref{tab:resnet_office_hom} since they do not release code.}. \vcode attains the state-of-the-art result.

\subsection{Results on Office-31}
We report the comparison with available baseline results on Office-31~\cite{saenko2010adapting} in \cref{tab:office_result}, using ResNet-34 backbone. Following \cite{saito2019semi}, two adaptation scenarios are compared (Webcam to Amazon, DSLR to Amazon). Our approach \code consistently outperforms the compared methods.

\subsection{Larger-shot results}
We provide \textit{10,20,50}-shot SSDA results on DomainNet in \cref{tab:domain-net_nshot}. We randomly select and add additional samples per class from the target domain to the target labeled pool. As a semi-supervised setting, we compared with both domain adaptation (DA) and semi-supervised learning (SSL) baselines~\cite{sohn2020fixmatch}. The implementation details are the same as those of \textit{1,3}-shot. \code improves along with more shots and can outperform baselines.

\subsection{Numbers and accuracy of pseudo-labels}
\begin{figure}[h]
    \vspace{-0.1in}
	\centerline{
    \includegraphics[width=0.9\linewidth]{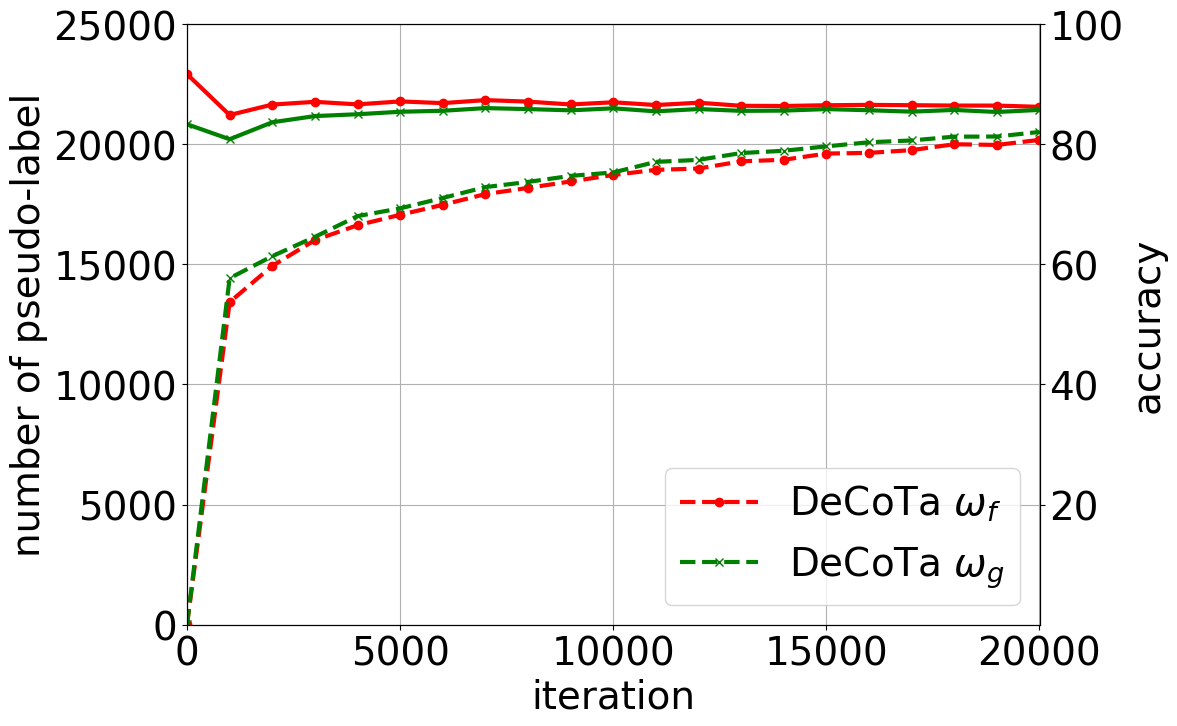}}
    \caption{\small Number (dashed, left) and accuracy (solid, right) of pseudo-labels on DomainNet three-shot setting, Real to Clipart.}
    \label{fig:pseudolabel_2classifiers}
\end{figure}

We showed the number of total and correct pseudo-labels by the two classifiers of \vcode along the training iterations in Figure 3 (c) of the main paper. The analysis is on DomainNet three-shot setting, from Real to Clipart. Concretely, for every $1K$ iterations (\ie, 24K unlabeled data), we accumulated the number of unlabeled data that have confident (with confidence $>\tau = 0.5$) and correct predictions by at least one classifier. We further plot them independently for each classifier (\ie, $\vw_{\color{red}f}$ and $\vw_{\color{blue}g}$) in \cref{fig:pseudolabel_2classifiers}. The accuracy of pseudo-labels remains stable (\ie, the number of confident and correct predictions divided by the number of confident predictions) but the number increases along training.

\subsection{Task decomposition}
We report the comparison of \vcode and \vmist on DomainNet and Office-Home in all the adaptation scenarios. As shown in \cref{tab:mist_mico}, \vcode outperform \vmist on all the setting by $1 \sim 2\%$ on DomainNet and $3 \sim 5\%$ on Office-Home, which further confirms the effectiveness of task decomposition --- explicitly considering the discrepancy between the two sources of supervision --- in \vcode.

\subsection{One-direction training}
We further consider another variant of \vcode named \textbf{one-direction teaching}, in which only one task teaches the other.
Instead of co-training, we use either $\vw_f$ or $\vw_g$ to generate pseudo-labels for both tasks\footnote{That is, \textbf{one-direction teaching} constructs both pseudo-label sets, \ie, $U^{(f)}$ and $U^{(g)}$ in \autoref{eq_code} of the main text, by the same model (we hence have two versions, $\vw_f$ teaching or $\vw_g$ teaching).}, while keeping the other setups the same as \vcode. 
This study is designed to measure the complementary specialties of the two tasks. 
As shown in \cref{tab:one_side_teaching}, the performance drops notably by using one-direction teaching. The results suggest that the two tasks provide unique expertise and complement each other, instead of one dominating the other.

\subsection{Results on the source domain}
We report the results on the source domain test set using $\vct{w}_f$ and $\vct{w}_g$ of \vcode on DomainNet (three-shot) in \cref{tab:source}.
While $\vw_f$ and $\vw_g$ have similar accuracy on the target domain test set, the fact that $\vw_f$ does not learn from $\DS$ suggests their difference in classifying source domain data. \cref{tab:source} confirms this: we see that $\vw_g$ clearly dominates $\vw_f$. Its accuracy is even on a par with a model trained only on $\DS$, showing one advantage of \code  --- the model can keep its discriminative ability on the source domain.

\subsection{Sensitivity to the confidence threshold $\tau$}
\begin{figure}[h]
    \vspace{-0.1in}
	\centerline{
    \includegraphics[width=0.9\linewidth]{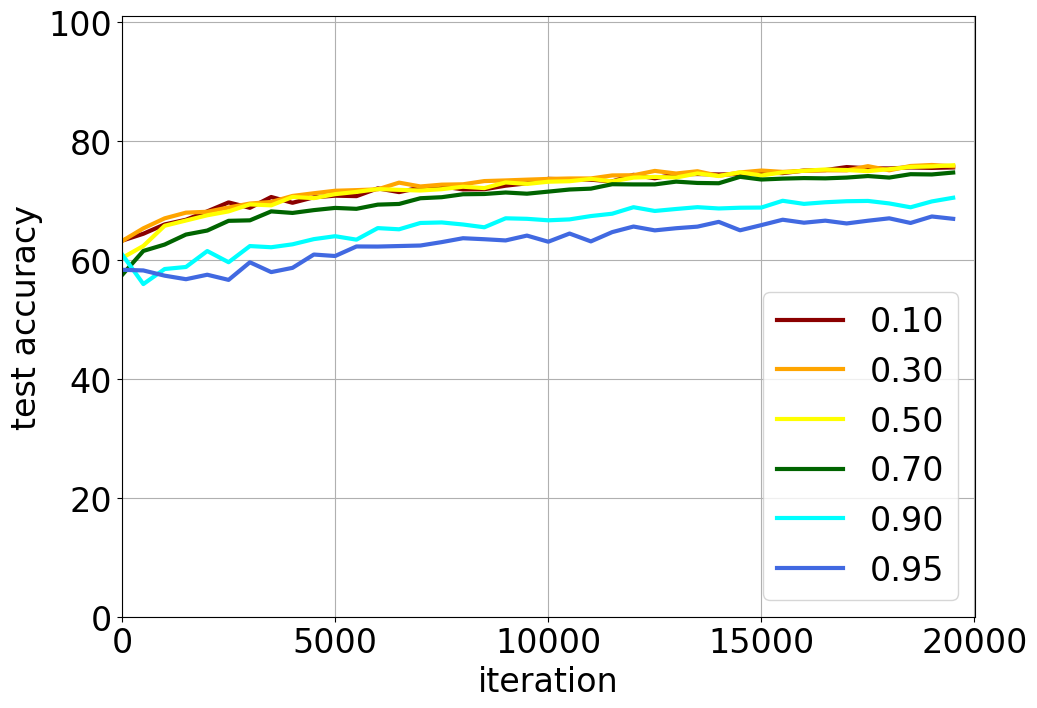}}
    \caption{\small \vcode's sensitivity to pseudo-label threshold $\tau$ on DomainNet three-shot setting, Real to Clipart.}
    \label{fig:tau_sensitivity}
\end{figure}

We investigate \vcode's sensitivity to the confidence threshold $\tau$ for assigning pseudo-labels (cf. \autoref{eq_code} and \autoref{eq_conf} of the main paper). 
As shown in \cref{fig:tau_sensitivity}, the variance in accuracy is small when $\tau\leq0.7$. The accuracy drops notably when $\tau\geq0.9$. 
We surmise that it is due to too few pseudo-labeled data are picked under a high threshold. 

\subsection{Analysis on the Beta distribution coefficient $\alpha$}
\begin{figure}[h]
	\vspace{-0.1in}
	\centerline{
    \includegraphics[width=0.9\linewidth]{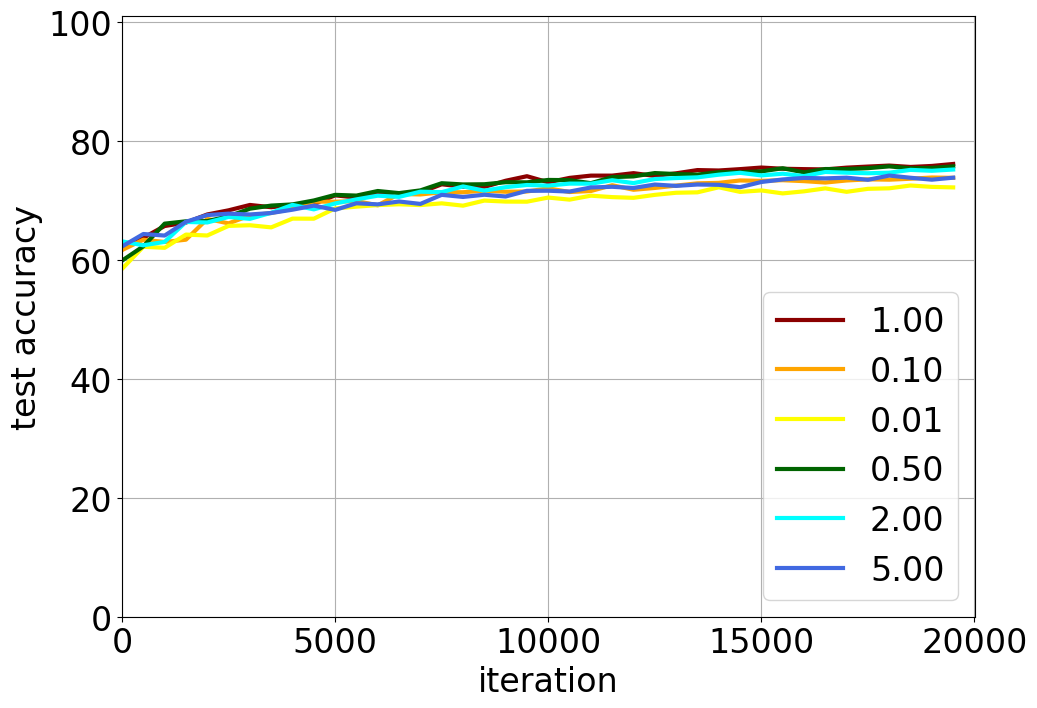}}
    \caption{\small \vcode's sensitivity to the Beta distribution coefficient $\alpha$ on DomainNet three-shot setting, Real to Clipart.}
    \label{fig:beta}
\end{figure}

\cref{fig:beta} shows \vcode's sensitivity to the \mixup hyper-parameter $\alpha$ in \autoref{eq_mixup} of the main paper: $\alpha$ is the coefficient of the Beta distribution, which influences the sampled value of $\lambda$, an indicator of the ``propotion'' in the \mixup algorithm. We report \vcode's result on DomainNet three-shot setting, adapting from Real to Clipart. The best performance is achieved by $\alpha=1.0$, equivalent to a uniform distribution of $\lambda\in [0, 1]$. This result is consistent with our hypothesis that \mixup connects the source and target domains with interpolated feature spaces in-between.

\begin{figure}[h]
    \vspace{-0.1in}
	\centerline{
    \includegraphics[width=0.8\linewidth]{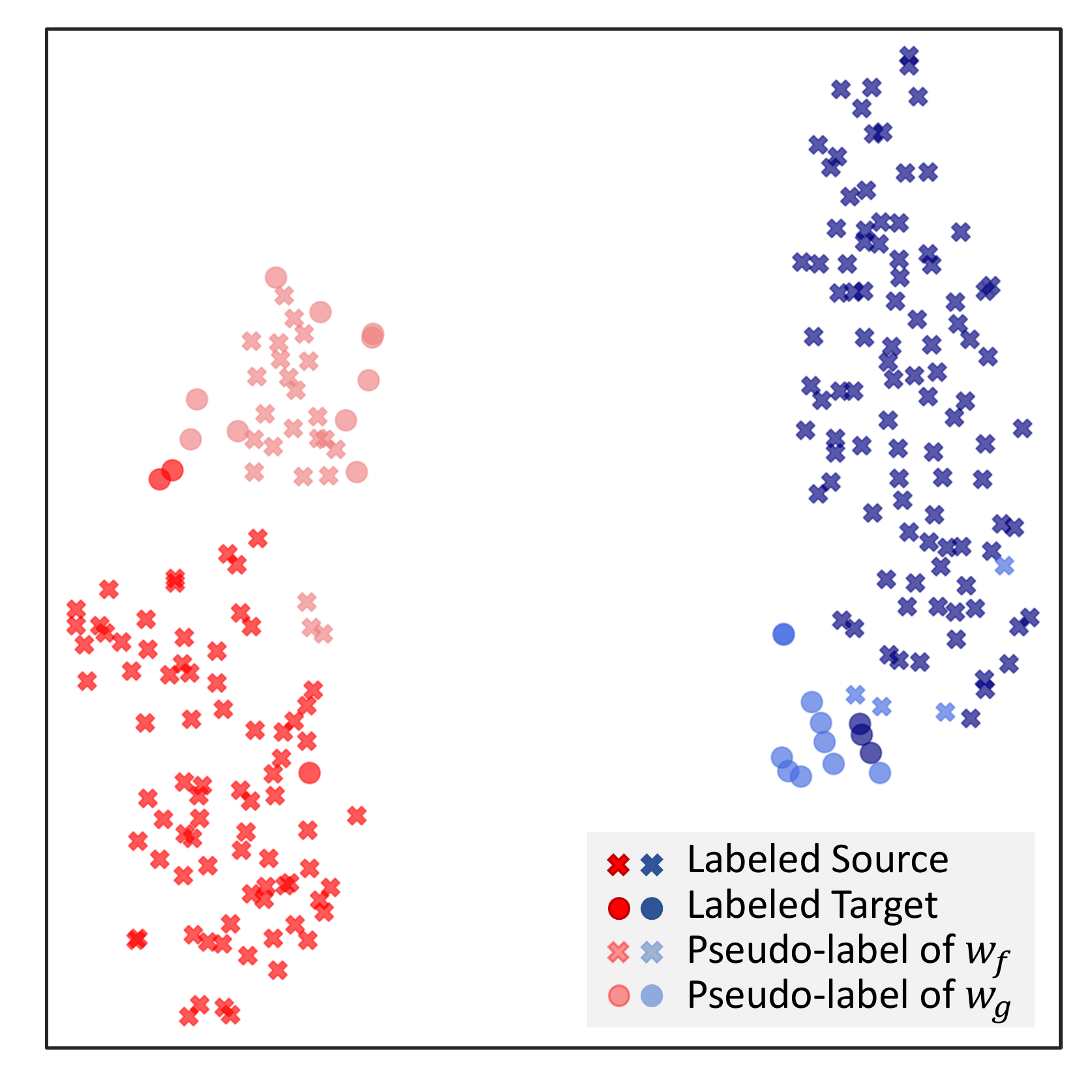}}
    \caption{\small t-SNE visualization of pseudo-labels assigned by $\vw_f$ and $\vw_g$ in \vcode (see text for details).}
    \label{fig:figure_1_tsne}
\end{figure}

\subsection{Training time} \code does not increase the training time much for two reasons. First, at each iteration (\ie, mini-batch), it only updates and learns from the pseudo-labels of \emph{the current mini-batch} of unlabeled data, not the entire unlabeled data. 
Second, assigning pseudo-labels only requires a forward pass of the mini-batch, just like most domain adaptation algorithms normally do to compute training losses.
The only difference is that \code trains two classifiers and needs to perform the forward pass of unlabeled data twice.

\subsection{t-SNE visualizations on \vcode tasks}
We visualize $\DS$, $\DT$, and the $\DU$ pseudo-labels by each task of \vcode in \cref{fig:figure_1_tsne}. For clarity, we select two classes for illustration. The colors blue and red represent the two classes; the shapes circle and cross represent data from $\DT$ (labeled target data) and $\DS$ (labeled source data), respectively. The colors light blue and light red represent the pseudo-labels of each class on $\DU$, 
in which the shape circle indicates that the pseudo-labels are provided by $\vw_f$ (learned with $\DT$) and the shape cross indicates that the pseudo-labels are provided by $\vw_g$ (learned with $\DS$).
The visualization is based on DomainNet three-shot setting, from Real to Clipart, trained for $10,000$ iterations. We see that $\vw_f$ tends to assign pseudo-labels to unlabeled data whose features are closer to $\DT$; $\vw_g$ tends to assign pseudo-labels to unlabeled data whose features are closer to $\DS$. Such a behavior is aligned with the seminal work of semi-supervised learning by~\cite{zhu2005semi}.

%
\begin{table*}[b!]
    \footnotesize
    \centering
    \renewcommand{\tabcolsep}{5pt}
    \renewcommand{\arraystretch}{1.0}
     \caption{\small Results on DomainNet at \textit{10, 20, 50}-shot, using ResNet-34. We tune hyper-parameters for SSL methods similarly to DA methods.
    }
    \resizebox{1.0\linewidth}{!}{
    \begin{tabular}{@{}l|ccc| ccc| ccc| ccc| ccc| ccc| ccc|ccc}
    \toprule
          & \multicolumn{3}{c}{R to C} & \multicolumn{3}{c|}{R to P} & \multicolumn{3}{c|}{P to C} & \multicolumn{3}{c|}{C to S} & \multicolumn{3}{c|}{S to P} & \multicolumn{3}{c|}{R to S} & \multicolumn{3}{c|}{P to R} & \multicolumn{3}{c}{Mean} \\
        \cmidrule(lr){2-4}
        \cmidrule(lr){5-7}
        \cmidrule(lr){8-10}
        \cmidrule(lr){11-13}
        \cmidrule(lr){14-16}
        \cmidrule(lr){17-19}
        \cmidrule(lr){20-22}
        \cmidrule(lr){23-25}
        n-shot $\rightarrow$ & 10 & 20 & 50 & 10 & 20 & 50 & 10 & 20 & 50 & 10 & 20 & 50 & 10 & 20 & 50 & 10 & 20 & 50 & 10 & 20 & 50 & 10 & 20 & 50 \\
        \midrule
          S+T & 69.1 & 72.4 & 77.5 & 67.3 & 70.2 & 73.4 & 68.2 & 72.5 & 77.7 & 62.9 & 67.3 & 71.8 & 64.8 & 67.9 & 72.6 & 61.3 & 65.5 & 70.2 & 78.0 & 79.3 & 82.2 & 67.4 & 70.7 & 75.1\\
         DANN~\cite{ganin2016domain} & 66.2 & 68.0 & 71.1 & 65.1 & 67.1 & 69.0 & 62.4 & 64.5 & 68.2 & 60.0 & 62.4 & 66.8 & 61.3 & 63.8 & 67.6 & 61.4 & 63.2 & 66.9 & 71.6 & 74.7 & 78.1 & 64.0 & 66.2 & 69.7 \\
        ENT~\cite{saito2019semi} & 77.9 & 80.0 & 83.0 & 72.3 & 74.9 & 77.7 & 77.5 & 79.1 & 82.3 & 66.3 & 70.1 & 75.0 & 66.3 & 71.0 & 75.7 & 63.9 & 68.3 & 74.6 & \textbf{81.2} & \textbf{82.9} & 84.5 & 72.2 & 75.2 & 79.0\\
        MME~\cite{saito2019semi} & 77.0 & 78.5 & 80.9 & 71.9 & 74.0 &76.4&75.6 &76.9 &80.4 &65.9 & 68.6 & 72.5 & 68.6 & 70.9 & 74.4 & 66.7 & 69.7 & 72.7 & 80.8 & 82.2 & 83.3 & 72.4 & 74.4 & 77.2 \\
        \midrule
        Mixup~\cite{zhang2017mixup} & 73.4 & 79.5 & 83.1 & 68.3 & 72.2 & 75.4 & 75.0 & 79.5 & 83.1 & 63.7 & 69.4 & 75.0 & 68.5 & 72.4 & 76.2 & 62.9 & 69.9 & 75.0 & 78.8 & 82.3 & \textbf{84.7} & 70.1 & 75.0 & 78.9\\
        FixMatch~\cite{sohn2020fixmatch}  & 76.6 & 79.5 & 82.3 & 73.0 & 74.7 & 76.4 & 75.8 & 79.4 & 83.3 & 70.1 & 73.1 & 76.9 & 71.3 & 73.3 & 77.0 & 68.7 & 71.6 & 74.2 & 79.7 & 81.9 & 84.2 & 73.6 & 76.2 & 79.2\\ 
        \midrule
        \vcode & \textbf{81.8} & \textbf{82.6} & \textbf{85.0} & \textbf{75.1} & \textbf{76.6} & \textbf{78.7} & \textbf{81.3} & \textbf{81.7} & \textbf{84.5} & \textbf{73.7} & \textbf{75.3} & \textbf{78.0} & \textbf{73.4} & \textbf{75.7} & \textbf{77.7} & \textbf{73.7} & \textbf{75.5} & \textbf{77.8} & 80.7 & 80.1 & 83.9 & \textbf{77.1} & \textbf{78.2} & \textbf{80.8}\\
         \bottomrule
    \end{tabular}
    }
    \label{tab:domain-net_nshot}
\end{table*}
%
\begin{table*}[]
\footnotesize
\centering
\caption{\small Comparison between \vcode and \vmist: test accuracy on DomainNet and Office-Home dataset ($\%$).}\vspace{-0.08in}
    \begin{subtable}{1.0\textwidth}
    \renewcommand{\tabcolsep}{6pt}
    \renewcommand{\arraystretch}{1.2}
    \centering
    \caption{DomainNet}\vspace{-0.05in}
        \begin{tabular}{@{}l|c|c c c c c c c|c@{}}
            \toprule
            Setting & Method & R to C & R to P & P to C & C to S & S to P & R to S & P to R & Mean \\
            \midrule
            \multirow{2}{*}{1-shot} & \vmist & 74.8 & 73.6 & 74.5 & 65.0 & 72.0 & 67.0 & 77.6 & 72.1\\
            & \vcode & 79.1 & 74.9 & 76.9 & 65.1 & 72.0 & 69.7 & 79.6 & 73.9\\
            \midrule
            \multirow{2}{*}{3-shot} &  \vmist & 78.1 & 75.2 & 76.7 & 68.3 & 72.6 & 71.5 & 79.8& 74.6\\
             & \vcode & 80.4 & 75.2 & 78.7 & 68.6 & 72.7 & 71.9 & 81.5 & 75.6\\
             \bottomrule
        \end{tabular}
        \vspace{0.05in}
    \end{subtable}
    \begin{subtable}{1.0\textwidth}
    \centering
    \renewcommand{\tabcolsep}{6pt}
    \renewcommand{\arraystretch}{1.2}
    \caption{Office-Home}\vspace{-0.05in}
    \begin{tabular}{@{}l|c|c c c c c c c c c c c c|c@{}}
    \toprule
    Setting & Method & R to C & R to P & R to A & P to R & P to C & P to A & A to P & A to C & A to R & C to R & C to A & C to P & Mean \\
    \midrule
    \multirow{2}{*}{1-shot} & \vmist & 42.7 & 77.5 & 62.9 & 73.1 & 39.4 & 54.8 & 67.1 & 40.0 & 66.9 & 67.9 & 56.8 & 69.4 & 59.9\\
    & \vcode & 47.2 & 80.3 & 64.6 & 75.5 & 47.2 & 56.6 & 71.1 & 42.5 & 73.1 & 71.0 & 57.8 & 72.9 & 63.3\\
    \midrule
    \multirow{2}{*}{3-shot} & \vmist & 54.7 & 81.2 & 64.0 & 69.4 & 51.7 & 58.8 & 69.1 & 47.6 & 70.6 & 65.3 & 60.8 & 73.8 & 63.9\\
    & \vcode & 59.9 & 83.9 & 67.7 & 77.3 & 57.7 & 60.7 & 78.0 & 54.9 & 76.0 & 74.3 & 63.2 & 78.4 & 69.3\\
    \bottomrule
     \end{tabular}
     \end{subtable}
\label{tab:mist_mico}
\end{table*}

\begin{table*}[]
    \footnotesize
    \renewcommand{\tabcolsep}{6pt}
    \renewcommand{\arraystretch}{1.2}
    \centering
    \caption{\small Comparison between \vcode and \textbf{one-direction teaching}: accuracy on DomainNet ($\%$) three-shot setting.}\vspace{-0.1in}
    \begin{tabular}{@{}l|c c c c c c c|c@{}}
        \toprule
        Method & R to C & R to P & P to C & C to S & S to P & R to S & P to R & Mean \\
        \midrule
         $\vct{w}_f$ teaching & 73.8 & 67.2 & 73.7 & 63.1 & 65.9 & 61.7 & 78.2 & 69.1 \\
         $\vct{w}_g$ teaching & 77.5 & 74.5 & 74.2 & 64.8 & 71.6 & 69.0 & 79.0 & 72.9 \\
         \vcode & 80.4 & 75.2 & 78.7 & 68.6 & 72.7 & 71.9 & 81.5 & 75.6\\
         \bottomrule
    \end{tabular}
    \label{tab:one_side_teaching}
\end{table*}

\begin{table*}[t]
    \footnotesize
    \renewcommand{\tabcolsep}{6pt}
    \renewcommand{\arraystretch}{1.2}
    \centering
    \caption{\small Comparison on the source domain test data of DomainNet ($\%$). Here we compare the two-task models of \vcode in the three-shot setting to the source-only model (S).}\vspace{-0.1in}
    \begin{tabular}{@{}l|c c c c c c c|c@{}}
        \toprule
        Method & R to C & R to P & P to C & C to S & S to P & R to S & P to R & Mean \\
        \midrule
         $\vct{w}_f$ & 55.2 & 68.2 & 43.8 & 59.5 & 50.8 & 56.9 & 61.0 & 56.3 \\
         $\vct{w}_g$ & 97.2 & 97.1 & 99.3 & 98.7 & 98.9 & 96.8 & 99.4 & 98.2 \\
         S & 98.1 & 98.2 & 99.5 & 98.9 & 99.2 & 98.2 & 99.6 & 98.8\\
         \bottomrule
    \end{tabular}
    \vskip5pt
    \label{tab:source}
\end{table*}

\end{document}